\documentclass[10pt,twocolumn,letterpaper]{article}

\usepackage[accsupp]{axessibility}  
\usepackage{wacv}
\usepackage{times}
\usepackage{epsfig}
\usepackage{graphicx}
\usepackage{amsmath}
\usepackage{amssymb}
\usepackage{booktabs}

\usepackage{latexsym}

\usepackage{bm}
\usepackage[ruled,vlined,linesnumbered]{algorithm2e}
\usepackage{multirow}
\usepackage{arydshln}
\usepackage{autobreak}
\usepackage{colortbl}
\usepackage{xurl}
\usepackage{soul}

\usepackage{tcolorbox}
\tcbuselibrary{theorems,skins}
\definecolor{mygray}{rgb}{0.9,0.9,0.9}

\def\wacvPaperID{1275} 

\wacvalgorithmstrack   

\wacvfinalcopy 

\ifwacvfinal
\usepackage[breaklinks=true,bookmarks=false,colorlinks]{hyperref}
\else
\usepackage[pagebackref=true,breaklinks=true,colorlinks,bookmarks=false]{hyperref}
\fi

\begin{document}

\title{Switching to Discriminative Image Captioning\\by Relieving a Bottleneck of Reinforcement Learning}

\author{
Ukyo Honda$^{1,2}$\qquad Taro Watanabe$^3$\qquad Yuji Matsumoto$^2$\\
$^1$CyberAgent, Inc.\qquad $^2$RIKEN\qquad $^3$Nara Institute of Science and Technology\qquad\\
\small{
\texttt{honda\_ukyo@cyberagent.co.jp}\qquad \texttt{taro@is.naist.jp}\qquad \texttt{yuji.matsumoto@riken.jp}
}
}

\maketitle
\thispagestyle{empty}

\begin{abstract}
Discriminativeness is a desirable feature of image captions: captions should describe the characteristic details of input images.
However, recent high-performing captioning models, which are trained with reinforcement learning (RL), tend to generate overly generic captions despite their high performance in various other criteria.
First, we investigate the cause of the unexpectedly low discriminativeness and show that RL has a deeply rooted side effect of limiting the output words to high-frequency words.
The limited vocabulary is a severe bottleneck for discriminativeness as it is difficult for a model to describe the details beyond its vocabulary.
Then, based on this identification of the bottleneck, we drastically recast discriminative image captioning as a much simpler task of encouraging low-frequency word generation.
Hinted by long-tail classification and debiasing methods, we propose methods that easily switch off-the-shelf RL models to discriminativeness-aware models with only a single-epoch fine-tuning on the part of the parameters.
Extensive experiments demonstrate that our methods significantly enhance the discriminativeness of off-the-shelf RL models and even outperform previous discriminativeness-aware methods with much smaller computational costs.
Detailed analysis and human evaluation also verify that our methods boost the discriminativeness without sacrificing the overall quality of captions.\footnote{The code is available at \url{https://github.com/ukyh/switch_disc_caption.git}}
\end{abstract}


\begin{figure}[t]
    \centering
    \includegraphics[width=1.0\columnwidth,keepaspectratio]{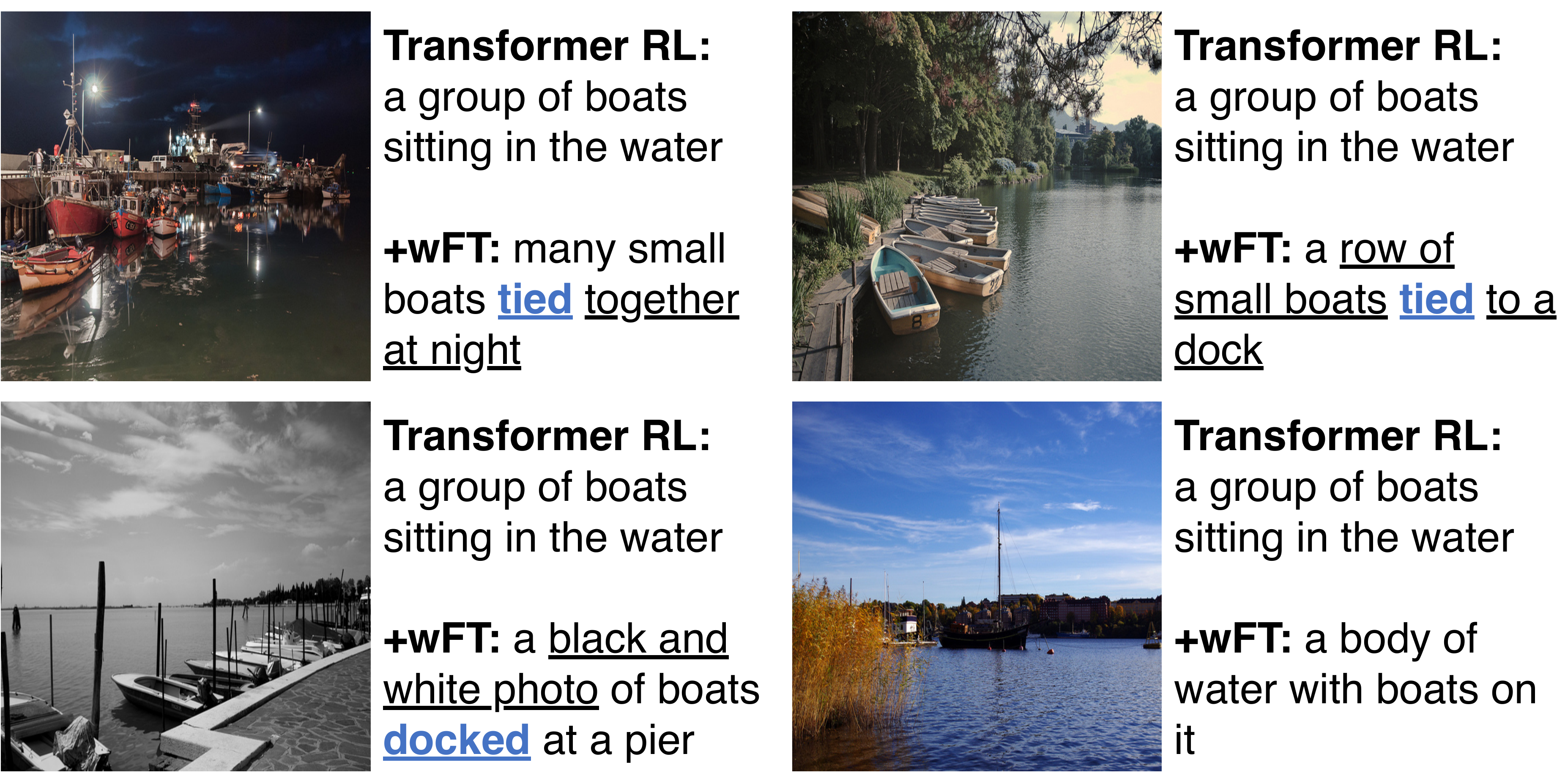}
    \caption{
    Caption examples in the MS~COCO validation set.
    \textbf{Transformer RL} is a Transformer captioning model trained with RL and \textbf{wFT} is our fine-tuning method.
    Transformer RL generates exactly the same caption for the four images. 
    The underlined words indicate the characteristic information that are not mentioned by Transformer RL, and the blue words are those that have never appeared in the outputs of the model.
    See Appendix~\textcolor{red}{2} for more examples.
    }
    \label{fig:headline}
\end{figure}

\section{Introduction}
Image captioning plays a fundamental role at the intersection of computer vision and natural language processing by converting the information in images into natural language descriptions. 
Generated captions can be used in various downstream tasks: aiding visually impaired users~\cite{gurari2020}, visual question answering on images and videos~\cite{fisch2020,kim2020dense}, visual dialogue~\cite{white2021open}, and news generation~\cite{zhang2021show}.

For those downstream tasks, captions should be \textbf{discriminative}: captions should describe the characteristic and important details of the input images~\cite{sadovnik2012image}.
However, current captioning models tend to generate overly generic captions~\cite{dai2017contrastive,dai2017towards,wang2019describing,wang2020towards}.
In particular, models trained with the standard \textbf{reinforcement learning (RL)}~\cite{rennie2017}, which is the \emph{de facto} standard training method in current image captioning~\cite{stefanini2021show}, unexpectedly perform poorly in discriminativeness despite the significant advantages in various other criteria~\cite{liu2019generating,wang2020compare}.
For example, a high-performing Transformer~\cite{vaswani2017attention} captioning model trained with RL generates exactly the same caption for the four different images shown in Figure~\ref{fig:headline}, ignoring the other salient details of each image.

To address the problem of overly generic captions, studies have been intensely conducted on \textbf{discriminative image captioning}, which is also called \textbf{distinctive} image captioning or \textbf{descriptive} image captioning.
Previous research has created new RL rewards regarding discriminativeness or new model architectures to enhance discriminativeness.
These approaches improved the discriminativeness; however, their models come with additional computations, require retraining from scratch, and do not shed light on the cause of \emph{existing models}' low discriminativeness.

Instead of creating or paying those computational costs, we first analyze the cause of the unexpectedly low discriminativeness of \emph{off-the-shelf RL models}, \emph{i.e.}, pre-trained, existing RL models, to explore ways to improve their discriminativeness.
\textbf{Our first contribution is the identification of a deeply rooted side effect in RL that limits output words to high-frequency words.}
The limited vocabulary is a severe bottleneck for discriminativeness as it is difficult for a model to describe the details beyond its vocabulary.

Given this identification of the bottleneck, now we can directly address the bottleneck by simply encouraging the generation of low-frequency words.
This task relaxation allows us to introduce long-tail classification and debiasing methods to discriminative image captioning for the first time.
\textbf{Our second contribution is our effective and efficient methods that switch any off-the-shelf RL models to discriminativeness-aware models with only a single-epoch fine-tuning on the part of the parameters.}
Unlike previous approaches, our methods do not require any discriminativeness rewards, new model architectures, or retraining from scratch.

Extensive experiments demonstrate that increasing low-frequency words in outputs significantly boosts discriminativeness from off-the-shelf RL models and even outperforms previous discriminativeness-aware models with much smaller computational costs.
These results verify that the limited vocabulary of RL models has been the major cause of their low discriminativeness.
Detailed analysis and human evaluation also show that our methods enhance the discriminativeness without sacrificing the overall quality.
We believe that our novel findings on the cause of low discriminativeness and the practical solutions to it will significantly impact future research on discriminative image captioning.

\section{Discriminativeness and a Bottleneck of RL}
Currently, RL is the \emph{de facto} standard training method for models used in image captioning because it significantly improves the performance in various evaluation metrics~\cite{stefanini2021show}.
However, it does not improve discriminativeness and may even decrease it~\cite{liu2019generating,wang2020compare}.
In this section, we examine the cause of the unexpectedly low discriminativeness.

\subsection{RL in Image Captioning}
\label{RL overview}
We provide a brief overview of the standard RL algorithm used in image captioning.
It was proposed by \cite{ranzato2015} and refined by \cite{rennie2017}.
Their goal was to directly optimize non-differentiable test-time metrics by minimizing the negative expected reward:
\begin{equation}
\label{eq:RL}
    \mathcal{L}_{\mathrm{RL}}(\theta)=-\mathbb{E}_{w^s \sim p_{\theta}(w^s \mid I)}[r(w^s)],
\end{equation}
where $w^s = (w^s_1,...,w^s_T)$ is a sequence sampled from a policy $p_{\theta}$, $I$ is an input image, and $r(\cdot)$ is a reward function.
To compute the gradient of $\mathcal{L}(\theta)$, \cite{ranzato2015} applied the REINFORCE algorithm~\cite{williams1992} to text generation.
The algorithm approximates the gradient as follows:
\begin{equation}
    \label{eq:RL grad}
    \nabla_{\theta}\mathcal{L}_{\mathrm{RL}}(\theta) \approx -(r(w^s) - b)\nabla_{\theta} \log p_{\theta}(w^s \mid I).
\end{equation}
Here, $b$ is a baseline reward that reduces the variance in the gradient.
Typically, the reward function $r(\cdot)$ is CIDEr~\cite{vedantam2015}, and the baseline reward $b$ is a reward for a sequence sampled with greedy decoding~\cite{rennie2017}.

\begin{figure}[t]
    \centering
    \includegraphics[width=1.0\columnwidth,keepaspectratio]{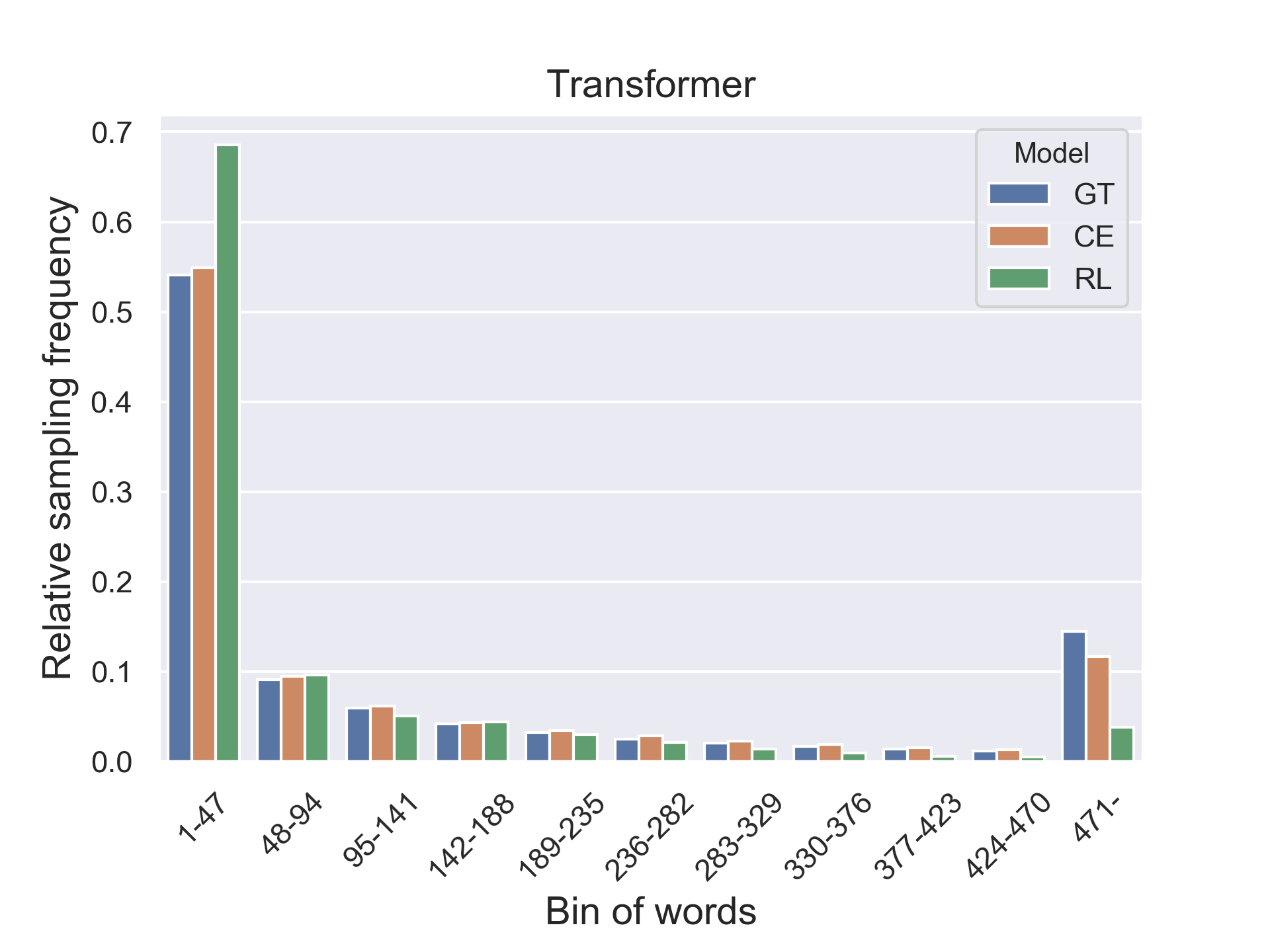}
    \caption{
    Relative frequency of the words in the sequences sampled for the MS~COCO training images.
    Five sequences were sampled for each image.
    The words (9,486 unique words excluding an out-of-vocabulary token $\langle \mathrm{unk} \rangle$) are sorted by their frequency in ground-truth captions and divided into 200 bins.
    We show the first 10 bins and the sum of the rest.
    GT is the ground-truth caption of the training images, CE is the output of a captioning model trained with the CE loss, and RL is the output of a captioning model trained with RL.
    Here, we used the Transformer model.
    }
    \label{fig:peakiness}
\end{figure}

\subsection{RL Limits Vocabulary}
\label{sec:RL to limited vocab}
Despite its effectiveness, RL has been found not to improve discriminativeness and somehow decrease the number of unique n-grams in output captions~\cite{liu2019generating, wang2020compare}.
As the relation between RL and these two negative effects is not obvious, it has been just considered a curious case.

We elucidate for the first time the relation between \emph{RL and limited vocabulary} by combining two recent findings.
\colorbox[rgb]{0.9,0.9,0.9}{\textbf{(1)}}~RL has been shown to make the output distribution peaky~\cite{choshen2019weaknesses,kiegeland2021revisiting}.
RL samples sequences from policy $p_{\theta}$ (See Eq.~\eqref{eq:RL}).
Typically, $p_{\theta}$ is initialized with a text-generation model pre-trained with the \textbf{Cross-Entropy (CE)} loss on ground-truth text.
In text generation, however, the initialized $p_{\theta}$ outputs peaky distributions, and thus, RL samples and rewards the words at the peak only, shaping more peaky distributions~\cite{choshen2019weaknesses}.
Then, where does $p_{\theta}$ tend to be peaky?
\colorbox[rgb]{0.9,0.9,0.9}{\textbf{(2)}}~Text-generation models have been theoretically and empirically shown to output distributions peaky at high-frequency words in the training corpus~\cite{nguyen2018improving,raunak-etal-2020-long,demeter2020stolen,holtzman2020curious}.
These two findings conclude that \emph{RL shifts the probability mass from low-frequency words to high-frequency words} by only sampling and rewarding the latter.

Figure~\ref{fig:peakiness} confirms the above by plotting the relative frequency of the words sampled for the training images.
The words are sorted by their frequency in ground-truth captions and divided into 200 bins.
Compared to the ground-truth captions and the sequences sampled with a CE model, the sequences sampled with an RL model are clearly limited to the high-frequency words, forming a peaky distribution\footnote{Although Figure~\ref{fig:peakiness} shows only the results obtained with the Transformer captioning model, we also confirmed that other models output peaky distributions~\cite{rennie2017,anderson2018}.
See Appendix~\textcolor{red}{3} for the details.}.

\subsection{Vocabulary Limits Discriminativeness}
Neural captioning models typically generate captions using sequential vocabulary-size classification~\cite{vinyals2015}.
However, the actual vocabulary a model can generate is much smaller than the entire vocabulary as the output distribution is highly skewed towards high-frequency words.
If the actual vocabulary cannot cover the details of an image, the model is forced to avoid those details and output only the information that high-frequency words can describe.
For example, the blue words in Figure~\ref{fig:headline} are not in the actual vocabulary of the RL model; these words have never been generated during evaluation.
As a result, the RL model had to ignore the characteristic relations \emph{tied} and \emph{docked} and ended up describing exactly the same for all four images.

Based on the observations, now we can hypothesize that the unexpectedly low discriminativeness of RL models has been rooted in the limited vocabulary.
This identification of the bottleneck is a key contribution as it allows us to address the low discriminativeness directly at the root.

\section{Methods to Relieve the Bottleneck}
\label{sec:method}
We have shown that RL results in the limited vocabulary as it steals the probability mass from low-frequency words.
Thus, increasing those low-frequency words is the easy yet critical solution to the bottleneck.
One way to achieve this is to jointly optimize both the RL loss and the CE loss on ground-truth captions so that the low-frequency words in ground-truth captions would be more likely to be sampled during RL training~\cite{wang2019describing}.
However, this approach still relies on the sampling from a skewed policy and requires retraining from scratch.

To increase the actual vocabulary more effectively and efficiently, we refine the mapping from encoded features to low-frequency words.
This refinement can be applied to any RL models and can be achieved by modifying only the mapping function parameters with a single-epoch fine-tuning.

\subsection{Simple Fine-Tuning (sFT)}
\label{sec:sFT}
The first method is a \textbf{simple fine-tuning (sFT)}.
It is based on a decoupled two-stage training~\cite{kang2019decoupling}, which is a current strong baseline model for long-tail classification~\cite{tang2020long,menon2020long,wang2020long}.
\cite{kang2019decoupling} decoupled the learning procedure into representation learning and classification, and then found that classification, \emph{i.e.}, the mapping from representations to label distributions, is critical for long-tail classification.
They decoupled the classification model $f_{\theta}(\cdot)$ into an encoder $g_{\theta_e}(\cdot)$ and a classifier consisting of weight and bias parameters: $f_{\theta}(x) = \bm{W}^\top g_{\theta_e}(x) + \bm{b}$.
Representation learning is the first stage of training, where they trained the entire classification model $f_{\theta}(\cdot)$ on a full training dataset.
The second stage is classification, where they fixed the encoder parameters $\theta_e$ and adjusted only the classifier parameters.
For the second-stage adjustment, they applied class-balanced sampling to encourage learning on low-frequency labels.

Following \cite{kang2019decoupling}, we decouple a captioning model into an encoder and a classifier.
In image captioning, the first-stage training of \cite{kang2019decoupling} corresponds to RL training on the full training dataset.
The second-stage training corresponds to adjusting the classifier parameters on the \emph{vocabulary-balanced} sequences.
However, sampling from the skewed policy of text-generation models cannot provide sequences containing low-frequency words (Section~\ref{sec:RL to limited vocab}).
Thus, we use ground-truth captions as relatively vocabulary-balanced samples.
sFT simply fine-tunes the classifier parameters of a pre-trained RL captioning model by minimizing the CE loss on ground-truth captions:
\begin{equation}
    \label{eq:sFT}
    \mathcal{L}_{\mathrm{CE}}(\hat{\theta}) = -\frac{1}{T} \sum^{T}_{t=1}{\log p_{\hat{\theta}}(w^g_t \mid w^g_{<t}, I)},
\end{equation}
where $w^g = (w^g_1, ..., w^g_T)$ is a ground-truth caption of image $I$ and \emph{$\hat{\theta}$ denotes the model parameters ${\theta}$ that are initialized with RL training}. 
Let the softmax function $\phi(\cdot)$ be
\begin{equation}
    \label{eq:softmax}
    \phi_{w_i,\beta}(\bm{z})=\frac{\exp(\beta z_{w_i})}{\sum_{w_j \in \mathcal{W}} \exp(\beta z_{w_j} )},
\end{equation}
where $z_{w_i}$ indicates the element of a vector $\bm{z} \in \mathbb{R}^{|\mathcal{W}|}$ at the index of a word $w_i \in \mathcal{W}$.
$\mathcal{W}$ is the entire vocabulary.
$\beta$ is an inverse-temperature hyperparameter that controls the steepness of the softmax distribution.
Then, the conditional probability $p_{\theta} (w^g_t \mid w^g_{<t}, I)$ is computed as follows:
\begin{align}
    \label{eq:prob}
    p_{\theta} (w^g_t \mid w^g_{<t}, I) &= \phi_{w^g_t,\beta}(s^t_{\theta}(w^g,I)),\\
    s^t_{\theta}(w^g,I) &= \bm{W}^{\top} g_{\theta_e} (w^g_{<t}, I) + \bm{b},
\end{align}
where $\bm{W} \in \mathbb{R}^{d \times |\mathcal{W}|}$ and $\bm{b} \in \mathbb{R}^{|\mathcal{W}|}$.
$d$ is the dimension of the hidden states of an encoder $g_{\theta_e}(\cdot)$.
We use LSTM~\cite{hochreiter1997} or Transformer~\cite{vaswani2017attention} for $g_{\theta_e}(\cdot)$.
During fine-tuning, \emph{only the classifier parameters $\{\bm{W}, \bm{b}\} \in \hat{\theta}$ are updated} with the gradients $\nabla_{\bm{W}} \mathcal{L}_{\mathrm{CE}}(\hat{\theta})$ and $\nabla_{\bm{b}} \mathcal{L}_{\mathrm{CE}}(\hat{\theta})$, respectively.

\subsection{Weighted Fine-Tuning (wFT)}
\label{sec:wFT}
Ground-truth captions contain more low-frequency words than sampled sequences, but some low-frequency words are still difficult to learn because of their low frequency.
Our second method is \textbf{weighted fine-tuning (wFT)}, which further pursues vocabulary balance by rebalancing the loss of high-frequency words and low-frequency words in ground-truth captions.

\begin{figure}[t]
    \centering
    \includegraphics[width=1.0\columnwidth,keepaspectratio]{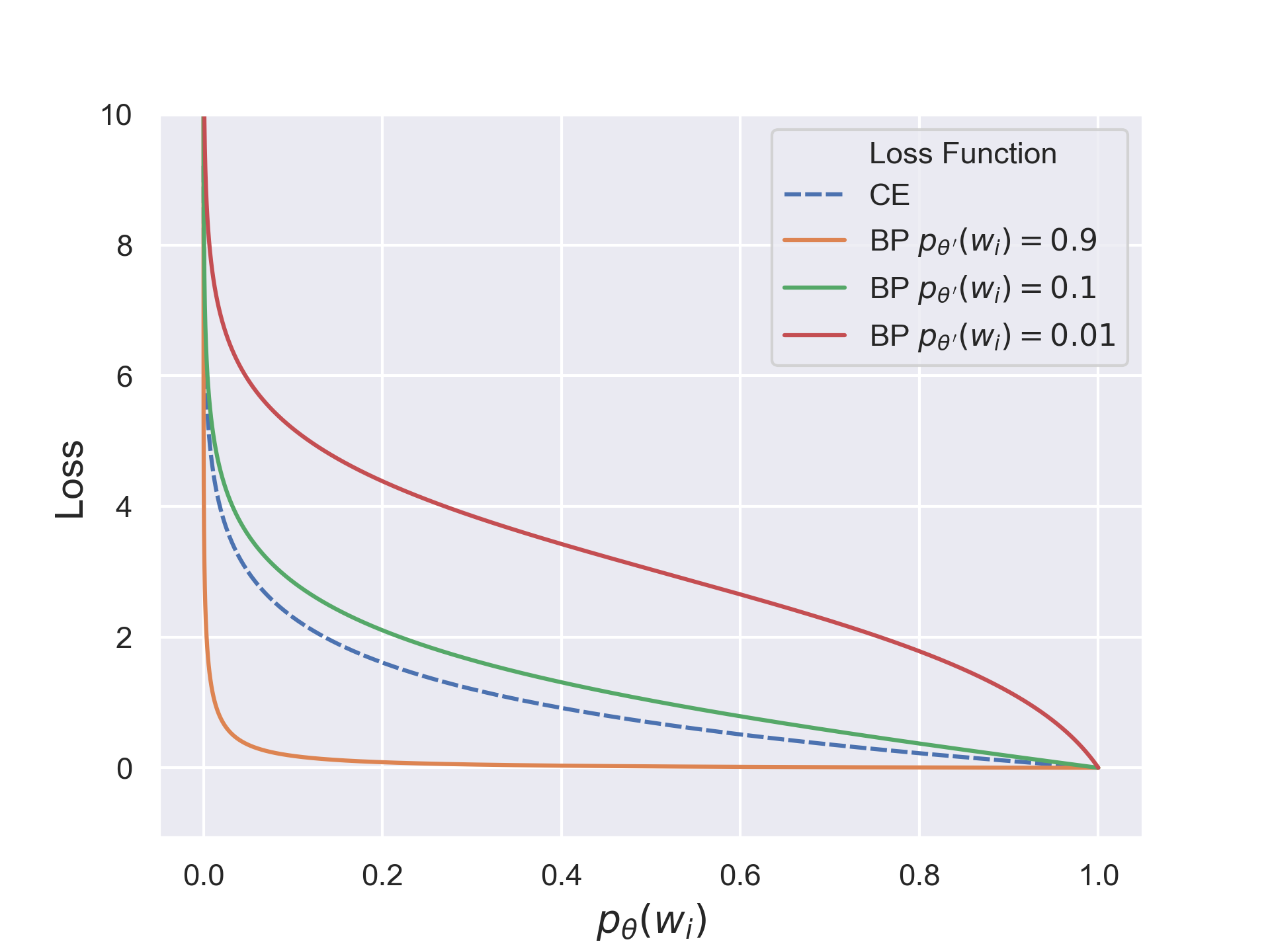}
    \caption{
    Visualization of the CE loss $-\log p_{\theta} (w_i)$ and BP loss $-\log p_{\theta, \theta'}(w_i)$.
    To compute the BP loss, we need the entire distribution of $\{p_{\theta}(w_i)\}_{w_i \in \mathcal{W}}$ and $\{p_{\theta'}(w_i)\}_{w_i \in \mathcal{W}}$.
    Here, we set the index $i$ to 1 and assigned $\frac{1}{5}(1 - p_{\theta}(w_1))$ to the words of the next five indices, $w_2, ..., w_6$.
    This is because we observed that the five most probable words occupied 99\% of the probability in the output distribution of the RL models.
    We assumed that the five most probable words were the same between $p_{\theta}$ and $p_{\theta'}$ as the parameters were initialized with the same RL model.
    Thus, we assigned $\frac{1}{5}(1 - p_{\theta'}(w_1))$ to the words of the next five indices, $w_2, ..., w_6$, likewise $p_{\theta}$.
    Here, $\beta$ and $\beta'$ were set to 1.
    }
    \label{fig:bp}
\end{figure}

To rebalance the loss, we exploit the frequency bias of RL models: RL models overly assign the probability to high-frequency words but not to low-frequency words.
Given the properties of the frequency bias, fine-tuning for discriminativeness should focus more on the words that an RL model is \emph{not} confident of but should be avoided on the words that an RL model is confident of.
wFT incorporates these heuristics by modifying the probability $p_{\theta}$ of $\mathcal{L}_{\mathrm{CE}}$ to the \textbf{bias product (BP)}~\cite{clark2019don,he2019unlearn,hinton2002training} probability, $p_{\theta, \theta'}$:
\begin{multline}
    \label{eq:BP}
    p_{\theta,\theta'}(w^g_t \mid w^g_{<t}, I) =\\
    \phi_{w^g_t,1}\Big[\log \tcboxmath[enhanced,frame hidden,colback=mygray,top=0pt,bottom=0pt,left=0pt,right=0pt,overlay={\node[below,black] at (frame.south) {\text{$p_{\theta}(\cdot \mid w^g_{<t}, I)$}};}]{\phi_{\cdot,\beta}(s^t_{\theta}(w^g,I))} + \log \tcboxmath[enhanced,frame hidden,colback=mygray,top=0pt,bottom=0pt,left=0pt,right=0pt,overlay={\node[below,black] at (frame.south) {\text{$p_{\theta'}(\cdot \mid w^g_{<t}, I)$}};}]{\phi_{\cdot,\beta'}(s^t_{\theta'}(w^g,I))}\Big],
\end{multline}
where $\phi_{\cdot,\beta}(\bm{z}) \in \mathbb{R}^{|\mathcal{W}|}$.
By inserting $p_{\theta, \theta'}$ into $\mathcal{L}_{\mathrm{CE}}$, we define the objective function of wFT as follows:
\begin{equation}
    \label{eq:wFT}
    \mathcal{L}_{\mathrm{BP}}(\hat{\theta}) = -\frac{1}{T} \sum^{T}_{t=1}{\log p_{\hat{\theta}, \hat{\theta}'}(w^g_t \mid w^g_{<t}, I)}.
\end{equation}
Similar to sFT, \emph{the parameters $\theta$ and $\theta'$ are initialized with the same RL model to be $\hat{\theta}$ and $\hat{\theta}'$}.
The difference is that, although the classifier parameters of $\hat{\theta}$ are updated, \emph{all the parameters of $\hat{\theta}'$ are fixed} during fine-tuning\footnote{\cite{chen2018factual} also utilized fixed pre-trained models to reweight their loss for stylized image captioning.
However, their method is designed to train new models from scratch and is not applicable to refining pre-trained models; their loss function (Eq.~(6) in \cite{chen2018factual}) is stuck at zero when we initialize the parameters with the same pre-trained model.
This requirement for retraining from scratch is a fundamental deviation from our goal of improving the discriminativeness of off-the-shelf RL models.
}.
Figure~\ref{fig:bp} shows the change in the BP loss compared to the CE loss.
The BP severely suppresses the loss when the frequency-biased policy $p_{\theta'}$ is confident, and largely increases the loss when $p_{\theta'}$ is not confident.
In this way, the BP allows models to unlearn the frequency bias learned with RL.
As with sFT, only the classifier parameters $\{\bm{W}, \bm{b}\} \in \hat{\theta}$ are updated with the gradients $\nabla_{\bm{W}} \mathcal{L}_{\mathrm{BP}}(\hat{\theta})$ and $\nabla_{\bm{b}} \mathcal{L}_{\mathrm{BP}}(\hat{\theta})$, respectively.

The previous BP methods used the probability $p_{\theta}$ during evaluation to avoid incorporating the bias of $p_{\theta'}$ into the predictions~\cite{clark2019don,he2019unlearn}.
Although it worked well in their classification tasks, we found this train--test gap makes the decoding unstable in text generation.
To mitigate the train--test gap, we use two variants of decoding: (1) decode with $p_{\theta}$ but use a small $\beta'$ for $p_{\theta'}$ during training to ease the gap between $p_{\theta}$ and $p_{\theta, \theta'}$, or (2) use $p_{\theta, \theta'}$ during both training and decoding (\textbf{BP decoding}) as $p_{\theta, \theta'}$ itself is already less biased than $p_{\theta'}$.

\begin{table*}[t]
\centering
\scalebox{0.70}{
\begin{tabular}{llcccccccccccc}  
\toprule
 &  & \multicolumn{3}{c}{\emph{Vocabulary}} & \multicolumn{6}{c}{\emph{Standard Evaluation}} & \multicolumn{3}{c}{\emph{Discriminativeness}} \\
\cmidrule(lr){3-5}
\cmidrule(lr){6-11}
\cmidrule(lr){12-14}
 &  & Unique-1 & Unique-S & Length & CIDEr & SPICE & BERTS+ & TIGEr & CLIPS & RefCLIPS & R@1 & R@5 & R@10 \\
\midrule
\multirow{9}*{\rotatebox[origin=c]{90}{\texttt{Att2in}}}
 & \textbf{Att2in RL} & 445 & 2,524 & 9.3 & 117.4 & 20.5 & 43.6 & 73.9 & 73.0 & 79.7 & 16.3 & 41.9 & 57.2 \\
 & \cellcolor[rgb]{0.9, 0.9, 0.9}\ \ + sFT & \cellcolor[rgb]{0.9, 0.9, 0.9}880 & \cellcolor[rgb]{0.9, 0.9, 0.9}3,156 & \cellcolor[rgb]{0.9, 0.9, 0.9}9.0 & \cellcolor[rgb]{0.9, 0.9, 0.9}115.4 & \cellcolor[rgb]{0.9, 0.9, 0.9}20.4 & \cellcolor[rgb]{0.9, 0.9, 0.9}\textbf{43.9} & \cellcolor[rgb]{0.9, 0.9, 0.9}74.3 & \cellcolor[rgb]{0.9, 0.9, 0.9}73.7 & \cellcolor[rgb]{0.9, 0.9, 0.9}\textbf{80.3} & \cellcolor[rgb]{0.9, 0.9, 0.9}20.1 & \cellcolor[rgb]{0.9, 0.9, 0.9}48.0 & \cellcolor[rgb]{0.9, 0.9, 0.9}62.8 \\
 & \cellcolor[rgb]{0.9, 0.9, 0.9}\ \ + wFT & \cellcolor[rgb]{0.9, 0.9, 0.9}\textbf{1,197} & \cellcolor[rgb]{0.9, 0.9, 0.9}\textbf{3,732} & \cellcolor[rgb]{0.9, 0.9, 0.9}8.9 & \cellcolor[rgb]{0.9, 0.9, 0.9}104.3 & \cellcolor[rgb]{0.9, 0.9, 0.9}19.5 & \cellcolor[rgb]{0.9, 0.9, 0.9}43.1 & \cellcolor[rgb]{0.9, 0.9, 0.9}74.2 & \cellcolor[rgb]{0.9, 0.9, 0.9}73.9 & \cellcolor[rgb]{0.9, 0.9, 0.9}80.2 & \cellcolor[rgb]{0.9, 0.9, 0.9}20.6 & \cellcolor[rgb]{0.9, 0.9, 0.9}49.7 & \cellcolor[rgb]{0.9, 0.9, 0.9}64.5 \\
 & \cellcolor[rgb]{0.9, 0.9, 0.9}\ \ + wFT (BP decoding) & \cellcolor[rgb]{0.9, 0.9, 0.9}1,102 & \cellcolor[rgb]{0.9, 0.9, 0.9}3,615 & \cellcolor[rgb]{0.9, 0.9, 0.9}9.4 & \cellcolor[rgb]{0.9, 0.9, 0.9}109.3 & \cellcolor[rgb]{0.9, 0.9, 0.9}20.1 & \cellcolor[rgb]{0.9, 0.9, 0.9}43.7 & \cellcolor[rgb]{0.9, 0.9, 0.9}\textbf{74.4} & \cellcolor[rgb]{0.9, 0.9, 0.9}\textbf{74.0} & \cellcolor[rgb]{0.9, 0.9, 0.9}80.2 & \cellcolor[rgb]{0.9, 0.9, 0.9}21.1 & \cellcolor[rgb]{0.9, 0.9, 0.9}\textbf{50.5} & \cellcolor[rgb]{0.9, 0.9, 0.9}64.8 \\
 & CIDErBtw & 470 & 2,630 & 9.3 & \textbf{119.0} & 20.7 & 43.8 & 74.1 & 73.1 & 79.8 & 17.2 & 44.1 & 58.7 \\
 & NLI & 465 & 2,626 & 9.2 & 118.9 & 20.6 & 43.8 & 74.1 & 73.2 & 79.9 & 17.6 & 44.4 & 59.8 \\
 & DiscCap$^{\dag}$ &  & 3,093 & 9.3 & 114.2 & \textbf{21.0} & &  &  &  & \textbf{21.6} & 50.3 & \textbf{65.4} \\
 & Joint CE & 700 & 2,907 & 9.1 & 111.7 & 19.9 & 43.5 & 74.0 & 73.3 & 80.0 & 19.1 & 46.7 & 61.5 \\
 & Only CE & 689 & 2,845 & 9.2 & 110.7 & 20.1 & 43.5 & 74.0 & 73.3 & 79.9 & 19.0 & 46.6 & 61.1 \\
\cdashline{1-14}
 & Visual Paraphrase$^{\dag}$ &  & 4,576 & 12.9 & 86.9 & 21.1 &  &  &  &  & 26.3 & 57.2 & 70.8 \\
\midrule
\multirow{8}*{\rotatebox[origin=c]{90}{\texttt{UpDown}}}
 & \textbf{UpDown RL} & 577 & 3,103 & 9.5 & 122.7 & \textbf{21.5} & 44.2 & 74.6 & 74.0 & 80.5 & 21.1 & 49.9 & 64.6 \\
 & \cellcolor[rgb]{0.9, 0.9, 0.9}\ \ + sFT & \cellcolor[rgb]{0.9, 0.9, 0.9}1,190 & \cellcolor[rgb]{0.9, 0.9, 0.9}3,788 & \cellcolor[rgb]{0.9, 0.9, 0.9}9.2 & \cellcolor[rgb]{0.9, 0.9, 0.9}115.9 & \cellcolor[rgb]{0.9, 0.9, 0.9}21.0 & \cellcolor[rgb]{0.9, 0.9, 0.9}44.2 & \cellcolor[rgb]{0.9, 0.9, 0.9}\textbf{74.9} & \cellcolor[rgb]{0.9, 0.9, 0.9}74.8 & \cellcolor[rgb]{0.9, 0.9, 0.9}\textbf{80.9} & \cellcolor[rgb]{0.9, 0.9, 0.9}25.0 & \cellcolor[rgb]{0.9, 0.9, 0.9}56.8 & \cellcolor[rgb]{0.9, 0.9, 0.9}71.2 \\
 & \cellcolor[rgb]{0.9, 0.9, 0.9}\ \ + wFT & \cellcolor[rgb]{0.9, 0.9, 0.9}\textbf{1,479} & \cellcolor[rgb]{0.9, 0.9, 0.9}\textbf{4,268} & \cellcolor[rgb]{0.9, 0.9, 0.9}9.1 & \cellcolor[rgb]{0.9, 0.9, 0.9}101.8 & \cellcolor[rgb]{0.9, 0.9, 0.9}19.5 & \cellcolor[rgb]{0.9, 0.9, 0.9}43.1 & \cellcolor[rgb]{0.9, 0.9, 0.9}74.6 & \cellcolor[rgb]{0.9, 0.9, 0.9}74.9 & \cellcolor[rgb]{0.9, 0.9, 0.9}80.7 & \cellcolor[rgb]{0.9, 0.9, 0.9}26.0 & \cellcolor[rgb]{0.9, 0.9, 0.9}57.6 & \cellcolor[rgb]{0.9, 0.9, 0.9}72.2 \\
 & \cellcolor[rgb]{0.9, 0.9, 0.9}\ \ + wFT (BP decoding) & \cellcolor[rgb]{0.9, 0.9, 0.9}1,275 & \cellcolor[rgb]{0.9, 0.9, 0.9}4,177 & \cellcolor[rgb]{0.9, 0.9, 0.9}9.6 & \cellcolor[rgb]{0.9, 0.9, 0.9}110.0 & \cellcolor[rgb]{0.9, 0.9, 0.9}20.6 & \cellcolor[rgb]{0.9, 0.9, 0.9}44.1 & \cellcolor[rgb]{0.9, 0.9, 0.9}\textbf{74.9} & \cellcolor[rgb]{0.9, 0.9, 0.9}\textbf{75.0} & \cellcolor[rgb]{0.9, 0.9, 0.9}80.8 & \cellcolor[rgb]{0.9, 0.9, 0.9}\textbf{26.7} & \cellcolor[rgb]{0.9, 0.9, 0.9}\textbf{58.7} & \cellcolor[rgb]{0.9, 0.9, 0.9}\textbf{72.4} \\
 & CIDErBtw & 582 & 3,108 & 9.4 & \textbf{123.0} & \textbf{21.5} & \textbf{44.4} & 74.6 & 74.2 & 80.7 & 21.9 & 50.9 & 65.9 \\
 & NLI & 575 & 3,144 & 9.4 & 122.4 & 21.4 & \textbf{44.4} & 74.6 & 74.1 & 80.6 & 21.5 & 50.7 & 65.6 \\
 & Joint CE & 857 & 3,120 & 9.4 & 111.8 & 20.5 & 43.7 & 74.3 & 73.8 & 80.2 & 21.8 & 51.2 & 65.2 \\
 & Only CE & 878 & 3,126 & 9.4 & 109.2 & 20.1 & 43.4 & 74.2 & 73.6 & 80.0 & 21.8 & 49.9 & 64.5 \\
\midrule
\multirow{8}*{\rotatebox[origin=c]{90}{\texttt{Transformer}}}
 & \textbf{Transformer RL} & 753 & 3,433 & 9.2 & 127.7 & 22.5 & 45.1 & 75.0 & 75.0 & 81.3 & 26.6 & 56.2 & 70.5 \\
 & \cellcolor[rgb]{0.9, 0.9, 0.9}\ \ + sFT & \cellcolor[rgb]{0.9, 0.9, 0.9}1,458 & \cellcolor[rgb]{0.9, 0.9, 0.9}3,959 & \cellcolor[rgb]{0.9, 0.9, 0.9}9.1 & \cellcolor[rgb]{0.9, 0.9, 0.9}118.7 & \cellcolor[rgb]{0.9, 0.9, 0.9}21.7 & \cellcolor[rgb]{0.9, 0.9, 0.9}44.8 & \cellcolor[rgb]{0.9, 0.9, 0.9}75.2 & \cellcolor[rgb]{0.9, 0.9, 0.9}75.6 & \cellcolor[rgb]{0.9, 0.9, 0.9}\textbf{81.5} & \cellcolor[rgb]{0.9, 0.9, 0.9}30.6 & \cellcolor[rgb]{0.9, 0.9, 0.9}62.3 & \cellcolor[rgb]{0.9, 0.9, 0.9}75.7 \\
 & \cellcolor[rgb]{0.9, 0.9, 0.9}\ \ + wFT & \cellcolor[rgb]{0.9, 0.9, 0.9}1,776 & \cellcolor[rgb]{0.9, 0.9, 0.9}4,274 & \cellcolor[rgb]{0.9, 0.9, 0.9}9.1 & \cellcolor[rgb]{0.9, 0.9, 0.9}103.1 & \cellcolor[rgb]{0.9, 0.9, 0.9}20.0 & \cellcolor[rgb]{0.9, 0.9, 0.9}43.3 & \cellcolor[rgb]{0.9, 0.9, 0.9}74.8 & \cellcolor[rgb]{0.9, 0.9, 0.9}75.8 & \cellcolor[rgb]{0.9, 0.9, 0.9}81.2 & \cellcolor[rgb]{0.9, 0.9, 0.9}32.5 & \cellcolor[rgb]{0.9, 0.9, 0.9}64.5 & \cellcolor[rgb]{0.9, 0.9, 0.9}77.1 \\
 & \cellcolor[rgb]{0.9, 0.9, 0.9}\ \ + wFT (BP decoding) & \cellcolor[rgb]{0.9, 0.9, 0.9}\textbf{1,964} & \cellcolor[rgb]{0.9, 0.9, 0.9}\textbf{4,373} & \cellcolor[rgb]{0.9, 0.9, 0.9}9.4 & \cellcolor[rgb]{0.9, 0.9, 0.9}107.3 & \cellcolor[rgb]{0.9, 0.9, 0.9}21.1 & \cellcolor[rgb]{0.9, 0.9, 0.9}44.2 & \cellcolor[rgb]{0.9, 0.9, 0.9}75.2 & \cellcolor[rgb]{0.9, 0.9, 0.9}\textbf{76.1} & \cellcolor[rgb]{0.9, 0.9, 0.9}\textbf{81.5} & \cellcolor[rgb]{0.9, 0.9, 0.9}\textbf{33.5} & \cellcolor[rgb]{0.9, 0.9, 0.9}\textbf{65.9} & \cellcolor[rgb]{0.9, 0.9, 0.9}\textbf{78.2} \\
 & CIDErBtw & 837 & 3,609 & 9.5 & 128.2 & 22.6 & 45.1 & 75.2 & 75.0 & 81.2 & 27.7 & 57.6 & 71.6 \\
 & NLI & 876 & 3,744 & 9.5 & \textbf{129.1} & \textbf{23.0} & \textbf{45.4} & \textbf{75.3} & 75.5 & \textbf{81.5} & 29.8 & 59.9 & 73.4 \\
 & Joint CE & 1,083 & 3,491 & 9.3 & 123.8 & 21.9 & 45.0 & 74.8 & 75.0 & 81.2 & 27.3 & 57.2 & 70.8 \\
 & Only CE & 935 & 3,599 & 9.4 & 112.2 & 20.8 & 44.0 & 74.5 & 74.8 & 80.9 & 26.5 & 55.8 & 69.7 \\
\bottomrule
\end{tabular}
}
\caption{
Comparison of \textbf{baseline models}, \colorbox[rgb]{0.9,0.9,0.9}{our models}, and state-of-the-art discriminativeness-aware models.
Automatic evaluation results on the MS~COCO test set.
\emph{Unique-1} and \emph{Unique-S} indicate the number of unique unigrams and sentences, respectively.
\emph{Length} is the average length of the output captions.
Scores with ${\dag}$ were reported in \cite{liu2019generating}.
Other scores were reproduced by us.
}
\label{tab:main}
\end{table*}

\section{Experiments}

\subsection{Setup}
\label{sec:setup}
\noindent \textbf{Dataset and Metrics.}
We used the MS~COCO captioning dataset\footnote{
Each split of training/validation/test contained 113,287/5,000/5,000 images, and each image had around five ground-truth captions.
}~\cite{lin2014mscoco,chen2015microsoft} with Karpathy splitting~\cite{karpathy2015}.
After preprocessing, the entire vocabulary size $|\mathcal{W}|$ was 9,487\footnote{The words that occur less than five times in the training captions were converted to $\langle \mathrm{unk} \rangle$ token.}.
In the evaluation, the captions were decoded using a beam search of size 5 and evaluated using various evaluation metrics.
Specifically, we used CIDEr~\cite{vedantam2015} SPICE~\cite{anderson2016}, Improved BERTScore (BERTS+)~\cite{yi2020improving}, TIGEr~\cite{jiang2019tiger}, CLIPScore (CLIPS), and RefCLIPScore (RefCLIPS)~\cite{hessel2021clipscore}.
Note that the correlation with human judgments increases in the above order, with RefCLIPS indicating the state-of-the-art correlation~\cite{hessel2021clipscore,kasai2022transparent}.
Following the previous studies~\cite{liu2019generating,wang2020compare,shi-etal-2021-enhancing}, we evaluated discriminativeness with \textbf{R@K} scores: the percentage of captions with which a pre-trained image--text retrieval model~\cite{faghri2018vsepp} could correctly retrieve the original images from the entire validation/test images within the rank of K $\in \{1, 5, 10\}$.
A higher R@K indicates that the model generates more discriminative captions with characteristic information of images.
Evaluation was conducted in a single run for each model.
See Appendix~\textcolor{red}{4} for the libraries and settings we used for these evaluations.

\noindent \textbf{Comparison Models.}
Following \cite{wang2020compare}, we used Att2in~\cite{rennie2017}, UpDown~\cite{anderson2018}, and Transformer~\cite{vaswani2017attention} as the baseline models.
The models were pre-trained with the standard RL~\cite{rennie2017} and are publicly available\footnote{\url{https://github.com/ruotianluo/self-critical.pytorch}: \texttt{\{Att2in, UpDown, Transformer\}+self\_critical}}.
In addition to the baseline models, we compared our models with state-of-the-art discriminativeness-aware models: \textbf{CIDErBtw}~\cite{wang2020compare}, \textbf{NLI}~\cite{shi-etal-2021-enhancing}, \textbf{DiscCap}~\cite{luo2018discriminability}, and \textbf{Visual Paraphrase}~\cite{liu2019generating}.
The first three created new discriminativeness rewards to be optimized with RL.
Visual Paraphrase introduced a new model architecture to paraphrase simpler captions to more complex captions.
See Section~\ref{sec:related work} for more details of these models.
As we mentioned in the beginning of Section~\ref{sec:method}, the CE loss on ground-truth captions can be utilized in a different way from our methods.
We report the results of jointly optimizing the RL loss and CE loss (\textbf{Joint CE}~\cite{wang2019describing,edunov2018classical}).
It optimizes $\mathcal{L}_\mathrm{Joint}(\theta) = \lambda \mathcal{L}_\mathrm{RL}(\theta) + (1 - \lambda) \mathcal{L}_\mathrm{CE}(\theta)$ during RL training.
We also tested \textbf{Only CE}, which sets $\lambda=0$ to solely optimize the CE loss, as the baseline without RL.
See Appendices~\textcolor{red}{12} and \textcolor{red}{13} for more comparisons~\cite{zhang2021vinvl,li2020oscar,cho-etal-2022-fine}.

\noindent \textbf{Hyperparameters.}
Our models used the same hyperparameters as the baseline models, except for the epoch size, learning rate, and $\beta'$ in Eq.~\eqref{eq:BP}.
We set the epoch size for fine-tuning to 1 and searched for the best learning rate from $\{1\textrm{e-}3, 1\textrm{e-}4, 1\textrm{e-}5, 1\textrm{e-}6\}$.
For BP in Eq.~\eqref{eq:BP}, we set $\beta = 1$ and searched for the best $\beta'$ from $\{0.1, 1\}$.
As with our models, we set all hyperparameters of the CE-based models to the same as the baseline models except for the $\lambda \in \{0, 0.2, 0.5, 0.8\}$.
We disabled scheduled sampling~\cite{bengio2015scheduled} for our fine-tuning and the CE loss to separate them from the RL loss strictly.
We took the best hyperparameters according to the R@1 scores in the validation set.
Note that \emph{we used different hyperparameters for the wFT with different decoding methods} (See Section~\ref{sec:wFT}).
Appendix~\textcolor{red}{5} shows the best hyperparameters.
We followed the previous work for the hyperparameters of the other models.

All the models except Visual Paraphrase had the same size of trainable parameters as their baselines.
See Appendix~\textcolor{red}{6} for the exact number of parameters.
Our fine-tuning was completed in around 10 minutes using a single GPU of 16 GB memory.
See Appendix~\textcolor{red}{7} for the exact time for training and comparison with other methods.

\subsection{Comparison with Baseline Models and Discriminativeness-Aware Models}
\label{sec:main exp}
Table~\ref{tab:main} shows the results compared to those obtained with the baseline models and state-of-the-art discriminativeness-aware models.

\noindent \textbf{Vocabulary.}
\label{sec:vocab exp}
First, we observe that our methods (sFT and wFT) successfully increase the actual vocabulary size: both of them considerably increased Unique-1 compared to all the baseline models.
wFT increased the vocabulary more than sFT, indicating that rebalancing the loss further encouraged low-frequency word generation.
The increased vocabulary resulted in more specific captions to each image: Unique-S also increased significantly.
Consistent with previous studies~\cite{wang2019describing,liu2019generating,wang2020compare}, the models trained with the CE loss (Joint CE and Only CE) achieved the larger vocabulary than the baseline RL models.
However, the improvement of our methods was even larger than these CE-based models.
Despite the significant increase in the vocabulary size, our method kept the captions concise: the average sentence length was close to those of the baseline models.

\noindent \textbf{Discriminativeness.}
\label{sec:discriminativeness exp}
Our goal is to enhance the discriminativeness of RL models by addressing their limited vocabulary.
As expected, our methods successfully improved the discriminativeness: the R@K scores of our models were considerably higher than those of the baselines.
Corresponding to the better improvement in vocabulary size, wFT increased discriminativeness more than sFT.
\emph{These results confirm our hypothesis that the limited vocabulary of RL models has been a major bottleneck for discriminativeness}.

Among the Att2in-based models, Visual Paraphrase achieved the highest discriminativeness.
However, this model is not directly comparable to the others because it increases the trainable parameters for its specialized model architecture.
Moreover, its improvement in discriminativeness was achieved at the expense of conciseness, which is another desirable property for discriminative image captions~\cite{sadovnik2012image}: its sentence length was substantially longer than the other models.
DiscCap performed comparably with our models, but its reward requires high computational costs.
CIDErBtw and NLI proposed more lightweight rewards to be applicable to larger models, but they still need retraining from scratch.
Among the larger models (UpDown and Transformer), our models achieved the highest discriminativeness despite the small computational cost.

\noindent \textbf{Standard Evaluation.}
\label{sec:standard exp}
As our methods increase low-frequency words in outputs, the outputs are likely to include the words that are \textbf{out-of-references (OOR)}.
That is, low-frequency words may not be covered by reference captions regardless of their correctness due to the low frequency.
These low-frequency OOR words unfairly decrease scores in conventional evaluation metrics because those metrics count \emph{exact matches} in the surface form of text\footnote{Some metrics use stemming, lemmatization, and/or WordNet synsets to evaluate synonyms but their coverage is limited.}.

To fairly evaluate the OOR words, recent metric research has focused on \emph{soft matching} metrics~\cite{jiang2019improving,hessel2021clipscore}.
Soft-matching metrics can evaluate the semantic similarity between target captions and reference captions beyond the surface form of text by utilizing pre-trained language models (PLMs)~\cite{yi2020improving,zhang2020bertscore} or pre-trained cross-modal models (PCMs)~\cite{jiang2019tiger,lee-etal-2020-vilbertscore,hessel2021clipscore}.
Their correlation with human judgments is significantly higher than that of exact-matching metrics in both precision and recall~\cite{kasai2022transparent}.
In particular, PCM-based metrics, which can utilize image features in addition to reference captions, have substantially enhanced the evaluation performance and have achieved the state-of-the-art correlation with human judgments~\cite{hessel2021clipscore,kasai2022transparent}.

Given the above advantages, we employed soft-matching metrics in addition to conventional exact-matching metrics.
Not surprisingly, our models decreased the scores in the exact-matching metrics (CIDEr and SPICE).
However, our models scored comparably with the baselines in the PLM-based metric (BERTS+) and rather outperformed them in the state-of-the-art PCM-based metrics (TIGEr, CLIPS, and RefCLIPS).
The higher performance in the superior soft-matching metrics indicates that our methods do not degrade the overall quality of captions.
To further validate the overall quality of our output captions, the following Section~\ref{sec:analysis} analyzes the cause of this performance gap in more detail.

\begin{table*}[t]
\centering
\scalebox{0.78}{
\begin{tabular}{lccccccccc}  
\toprule
 & \multicolumn{3}{c}{} & \multicolumn{3}{c}{\emph{Text-Based}} & \multicolumn{3}{c}{\emph{Text-and-Image-Based}} \\
\cmidrule(lr){5-7}
\cmidrule(lr){8-10}
 & \emph{Repetition} & \multicolumn{2}{c}{\emph{OOR}} & \multicolumn{2}{c}{\emph{Exact-Matching}} & \multicolumn{4}{c}{\emph{Soft-Matching}} \\
\cmidrule(lr){2-2}
\cmidrule(lr){3-4}
\cmidrule(lr){5-6}
\cmidrule(lr){7-10}
 & Rep (\%)~$\downarrow$ & Number~$\downarrow$ & Rank~$\uparrow$ & CIDEr & SPICE & BERTS+ & TIGEr & CLIPS & RefCLIPS \\
\midrule
\textbf{Att2in RL} & 4.1 & \textbf{8,665} & 79.4 & \textbf{117.4} & \textbf{20.5} & 43.6 & 73.9 & 73.0 & 79.7 \\
\rowcolor[rgb]{0.9, 0.9, 0.9}
\ \ + sFT & 3.8 & 8,813 & 164.0 & 115.4 & 20.4 & \textbf{43.9} & 74.3 & 73.7 & \textbf{80.3} \\
\rowcolor[rgb]{0.9, 0.9, 0.9}
\ \ + wFT & \textbf{3.2} & 10.454 & \textbf{237.9} & 104.3 & 19.5 & 43.1 & 74.2 & 73.9 & 80.2 \\
\rowcolor[rgb]{0.9, 0.9, 0.9}
\ \ + wFT (BP decoding) & 3.6 & 10,386 & 204.7 & 109.3 & 20.1 & 43.7 & \textbf{74.4} & \textbf{74.0} & 80.2 \\
Only CE & 3.9 & 9,913 & 133.1 & 110.7 & 20.1 & 43.5 & 74.0 & 73.3 & 79.9 \\
\midrule
\textbf{UpDown RL} & 3.9 & \textbf{8,463} & 100.1 & \textbf{122.7} & \textbf{21.5} & \textbf{44.2} & 74.6 & 74.0 & 80.5 \\
\rowcolor[rgb]{0.9, 0.9, 0.9}
\ \ + sFT & 3.6 & 9,252 & 225.8 & 115.9 & 21.0 & \textbf{44.2} & \textbf{74.9} & 74.8 & \textbf{80.9} \\
\rowcolor[rgb]{0.9, 0.9, 0.9}
\ \ + wFT & \textbf{3.0} & 11,478 & \textbf{301.0} & 101.8 & 19.5 & 43.1 & 74.6 & 74.9 & 80.7 \\
\rowcolor[rgb]{0.9, 0.9, 0.9}
\ \ + wFT (BP decoding) & 3.4 & 11,065 & 236.9 & 110.0 & 20.6 & 44.1 & \textbf{74.9} & \textbf{75.0} & 80.8 \\
Only CE & 3.7 & 10,874 & 152.9 & 109.2 & 20.1 & 43.4 & 74.2 & 73.6 & 80.0 \\
\midrule
\textbf{Transformer RL} & 3.6 & \textbf{7,824} & 129.8 & \textbf{127.7} & \textbf{22.5} & \textbf{45.1} & 75.0 & 75.0 & 81.3 \\
\rowcolor[rgb]{0.9, 0.9, 0.9}
\ \ + sFT & 3.2 & 9,397 & 296.0 & 118.7 & 21.7 & 44.8 & \textbf{75.2} & 75.6 & \textbf{81.5} \\
\rowcolor[rgb]{0.9, 0.9, 0.9}
\ \ + wFT & \textbf{2.6} & 11,930 & 379.7 & 103.1 & 20.0 & 43.3 & 74.8 & 75.8 & 81.2 \\
\rowcolor[rgb]{0.9, 0.9, 0.9}
\ \ + wFT (BP decoding) & 2.9 & 11,673 & \textbf{461.0} & 107.3 & 21.1 & 44.2 & \textbf{75.2} & \textbf{76.1} & \textbf{81.5} \\
Only CE & 3.3 & 10,661 & 165.6 & 112.2 & 20.8 & 44.0 & 74.5 & 74.8 & 80.9 \\
\midrule
\midrule
Human & \textbf{2.4} & 17,963 & \textbf{815.6} & 88.4 & 21.2 & 42.9 & 73.3 & \textbf{77.7} & \textbf{82.0} \\
\bottomrule
\end{tabular}
}
\caption{
Comparison of OOR words and the resulting difference in exact-matching and soft-matching metrics.
We report the results on the MS~COCO test set.
A higher value in \emph{Rank} indicates a lower frequency rank of the OOR words.
We also report the rate of repetition.
}
\label{tab:text-based analysis}
\end{table*}

\begin{table}[t]
\centering
\scalebox{0.77}{
\begin{tabular}{llll}  
\toprule
 & Discriminativeness & Correctness & Fluency \\
\midrule
\textbf{Transformer RL} & \underline{3.00} & 4.42 & 4.83 \\
\rowcolor[rgb]{0.9, 0.9, 0.9}
\ \ + wFT & \textbf{3.34}$^{**}$ & 4.45 & \textbf{4.84} \\
NLI & 3.18$^{**}$ & \textbf{4.54} & 4.76 \\
\bottomrule
\end{tabular}
}
\caption{Human evaluation results on the subset of the MS~COCO test set.
The discriminativeness score of Transformer RL was fixed at 3.00 because we set it as the baseline.
$^{*}$/$^{**}$ indicates that a score is statistically significantly different from that of the baseline model (t-test with $p < 0.05/0.01$); one-sample t-test for discriminativeness and independent two-sample t-test for the other criteria.
}
\label{tab:human eval}
\end{table}

\subsection{Analysis of the Performance Gap}
\label{sec:analysis}
\noindent \textbf{Properties of OOR Words.}
The critical difference between the conventional exact-matching metrics and the recent soft-matching metrics is the (in)ability to evaluate OOR words\footnote{Note that this difference does not mean that exact-matching metrics represent precision, and soft-matching metrics represent recall.
Exact-matching metrics cannot represent precision because the reference captions do not cover all correct descriptions.
That is, exact-matching metrics can only represent the flawed precision with false negatives.
Actually, exact-matching metrics correlate with human judgments worse than soft-matching metrics not only in recall but also in precision~\cite{kasai2022transparent}.}.
Based on the difference, we hypothesize that the performance gap is caused by a difference in the properties of OOR words.
We analyzed the OOR words of our models, comparing with those of RL baselines and Only CE, which scores similarly to our models in exact-matching metrics but decreases soft-matching scores in contrast to our models.
Table~\ref{tab:text-based analysis} shows the number of OOR words and their average \textbf{frequency rank}.
The frequency rank refers to the order of words when sorted by their frequency in training captions; the most frequent word ranks 1st, and the value of rank increases as the frequency decreases.
Although our models and Only CE output the similar number of OOR words, the significant difference in the frequency rank indicates that the properties of our OOR words are different from those of Only CE; that is, the OOR words of our models consist of much more low-frequency words than those of Only CE.
Low-frequency words are likely to be OOR by the nature of their frequency, regardless of their correctness.

The soft-matching metrics could tell this difference and scored our models higher than Only CE models and even higher than baseline RL models.
Especially, this tendency was more clear in the state-of-the-art PCM-based metrics (TIGEr, CLIPS, and RefCLIPS).
On the contrary, the exact-matching metrics (CIDEr and SPICE) could not tell the difference by definition and decreased the scores roughly in proportion to the number of OOR words.
Appendix~\textcolor{red}{8} shows the qualitative analysis of the underrated captions.

\noindent \textbf{Comparison with Human-Annotated Captions.}
Human-annotated captions are known to show low exact-matching scores despite their high quality~\cite{kasai2022transparent,liu2019generating,dai2017towards}.
In Table~\ref{tab:text-based analysis}, we observe that human-annotated captions (\emph{Human})\footnote{Following~\cite{liu2019generating,dai2017towards}, we randomly sampled one reference caption for each image and evaluated the similarity against the rest of the references.} have similar properties to ours: a large number of low-frequency OOR words, low exact-matching scores, but high scores in the state-of-the-art metrics (CLIPS and RefCLIPS).

\noindent \textbf{Repetition.}
We also confirmed that the decrease in exact-matching scores was not caused by repetition, which is a typical side effect of heavily maximizing discriminativeness rewards~\cite{wang2019describing,vered2019joint}.
Table~\ref{tab:text-based analysis} shows that our models' repetition rates\footnote{Let $\mathcal{C}$ be a set of captions; $f^n(\cdot)$ and $u^n(\cdot)$ be the functions to return $n$-grams and unique $n$-grams, respectively. We computed the repetition rate (\emph{Rep}) by $\frac{1}{|\mathcal{C}|N} \sum_{i=1}^{|\mathcal{C}|} \sum_{n=1}^{N} 1 - \frac{|u^n(\mathcal{C}_i)|}{|f^n(\mathcal{C}_i)|}$, where we set $N=4$.} were rather lower than those of baselines.

\noindent \textbf{Conclusion.}
From the above results, we conclude that the lower exact-matching scores of our models are caused by the nature of low-frequency words and the deficiency of exact-matching metrics, not by the degeneration of our models.
The results of the human evaluation in the following Section~\ref{sec:human eval} further support this conclusion.

\subsection{Human Evaluation}
\label{sec:human eval}
As discussed in Sections~\ref{sec:standard exp}~and~\ref{sec:analysis}, automatic evaluation of our models has difficulty due to the OOR words caused by the low frequency.
To further validate the performance of our models, we conducted human evaluations using Amazon Mechanical Turk (AMT) on three criteria: discriminativeness, correctness, and fluency.
Correctness and fluency are \emph{absolute scores}: we instructed workers to give a maximum score 5 to the captions that \emph{did not} contain incorrect information (ungrammatical or unnatural expressions) in terms of correctness (fluency).
In contrast, discriminativeness is designed as a \emph{relative score} because it is difficult to set an absolute standard for discriminativeness; unlike correctness or fluency, we cannot define the perfectly discriminative captions.
Following \cite{wang2020compare}, we instructed the workers to determine the discriminativeness of a caption by comparing the caption with that of a baseline model\footnote{If a target caption describes the same information as a baseline caption, the workers give the target caption a score of 3; if the target caption describes more (less) characteristic information than the baseline caption, the workers give the target caption a score of 4 or 5 (1 or 2).}.

We evaluated the Transformer-based models, which performed the best in the automatic evaluation.
Although wFT with BP decoding performed better, here we picked up wFT with $p_{\theta}$ decoding to set the total number of parameters for decoding strictly the same across the models.
Following \cite{wang2020compare}, we randomly selected 50 images from the MS~COCO test set and assigned five workers to each image.
See Appendix~\textcolor{red}{9} for more details on the AMT instruction.
Table~\ref{tab:human eval} shows the results.
wFT, which had the highest R@K scores, also achieved the highest discriminativeness here.
wFT achieved the same or higher correctness and fluency than the baseline model, in contrast to the exact-matching scores in Table~\ref{tab:text-based analysis}.
These results are consistent with the results of the state-of-the-art soft-matching metrics, confirming again that our methods do not degrade the quality of captions.

\section{Related Work}
\label{sec:related work}
\noindent \textbf{Image Captioning} is the task of describing images in natural languages.
The quality of captions has been remarkably improved by recent advances such as the encoder--decoder captioning model~\cite{vinyals2015}, attention mechanism~\cite{xu2015}, RL training~\cite{ranzato2015,rennie2017}, attention over bounding box features~\cite{anderson2018}, large-scale pre-training~\cite{li2020oscar}, and large-scale captioning datasets~\cite{young2014image,lin2014mscoco,chen2015microsoft,krishna2017,sharma2018}.
Despite these advancements, current captioning models generate overly generic captions~\cite{dai2017contrastive,dai2017towards,wang2019describing,wang2020towards}.

\noindent \textbf{Discriminative Image Captioning} has been explored to generate more informative captions.
\cite{sadovnik2012image} was the first to study it.
They defined the more informative captions as the captions that \emph{concisely} describe the information discriminative from \emph{distractor images}, \emph{i.e.}, images similar to an input image.
\cite{andreas2016reasoning} proposed neural listener and speaker models that cooperate to generate discriminative captions for abstract scenes.
\cite{monroe2017colors} adapted the models to single-colored images.
\cite{vedantam2017context,cohn2018pragmatically} extended the domain to real images and improved inference efficiency.
\cite{wang2021group} proposed a memory attention network to describe unique objects among distractor images.
\cite{mao2022rethinking} introduced a dataset with harder distractor images.

These approaches require selecting distractor images for inference.
\cite{luo2018discriminability} and \cite{liu2018show} proposed the methods that do not require this step. 
Their models learn to generate discriminative captions by maximizing the R@K scores for sampled captions using RL~\cite{rennie2017}.
The R@K scores are computed with a pre-trained image--text retrieval model~\cite{faghri2018vsepp} over images in a mini-batch.
\cite{vered2019joint} proposed a method to jointly train the image--text retrieval model and captioning model.
Despite their effectiveness, R@K scores are associated with high computational costs and require a large batch size.
Recently, CIDErBtw~\cite{wang2020compare} and NLI~\cite{shi-etal-2021-enhancing} achieved state-of-the-art discriminativeness with more lightweight rewards.
They weighted the contribution of ground-truth captions for the CIDEr reward according to their differences from similar but different captions~\cite{wang2020compare} or their entailment scores against other ground-truth captions~\cite{shi-etal-2021-enhancing}.
Another approach exploited unrelated captions as negative examples and trained caption generators with contrastive learning~\cite{dai2017contrastive} or GAN~\cite{dai2017contrastive,goodfellow2014}.

Visual Paraphrase~\cite{liu2019generating} and \cite{wu2020fine} are related to our work in that they exploited low-frequency n-grams to enhance discriminativeness.
\cite{liu2019generating} divided ground-truth captions into two subsets according to n-gram TF-IDF scores and proposed a new model to paraphrase low TF-IDF captions into high TF-IDF ones.
\cite{wu2020fine} proposed the use of n-gram TF-IDF scores as an additional reward to a variant of R@K reward.

Different from above approaches, our objective is set to remedy the low discriminativeness of existing RL models.
Our models can be achieved with single-epoch fine-tuning of pre-trained RL models, without requiring either drastic changes in the model architecture~\cite{liu2019generating}, additional computational costs of rewards~\cite{wu2020fine}, or retraining from scratch.

\noindent \textbf{Diverse Image Captioning} is the task of generating a set of diverse captions for a given image~\cite{wang2016diverse}.
Diverse image captioning is aimed at enumerating various pieces of information with a set of captions, whereas discriminative image captioning aims to concisely describe the most characteristic information with a single caption.
Similar to this study, some studies utilized captions that contained more low-frequency words, such as ground-truth captions~\cite{wang2019describing,luo2020analysis} or captions sampled from CE models~\cite{shi2021partial}.
Their models learn to generate these captions in addition to the captions sampled from RL models.
However, these approaches still rely on sampling from skewed policies and require retraining of a model from scratch.

\noindent \textbf{Long-Tail Classification} has been studied extensively in various tasks as label imbalance is prevalent across datasets~\cite{zhang2021deep,li2020dice}.
In text-generation tasks, label imbalance exists in the frequency of words.
Previous approaches have addressed the imbalance by normalizing classifier weights~\cite{nguyen2018improving,raunak-etal-2020-long} or using variants of Focal loss~\cite{raunak-etal-2020-long,gu2020token,jiang2019improving,wu2020importance,lin2017focal}.
In contrast to these approaches, we adapted long-tail classification to mitigate the side effects of RL in the context of discriminative image captioning.
Appendix~\textcolor{red}{10} shows that our methods outperformed these approaches.

\section{Conclusion}
We have investigated the cause of overly generic captions of RL models and found out that RL decreases the discriminativeness by limiting the output words to high-frequency words.
We propose the lightweight fine-tuning methods to address the bottleneck directly and achieve significantly higher discriminativeness with only the slight modification on off-the-shelf RL models.
Our identification of the bottleneck and practical solutions will significantly impact future research on discriminative image captioning.

As an additional practical advantage, our models can control the granularity of descriptions from coarse to fine by just switching the off-the-shelf/fine-tuned classifier parameters.
In terms of broader impact, our methods can be easily applied to the RL models in other text generation tasks, such as machine translation~\cite{wu-etal-2018-study}, summarization~\cite{pasunuru-bansal-2018-multi}, and dialogue generation~\cite{li-etal-2016-deep} to enrich the output vocabulary.

{\small
\bibliographystyle{ieee_fullname}
\bibliography{wacv2023}

\begin{thebibliography}{10}\itemsep=-1pt

\bibitem{anderson2016}
Peter Anderson, Basura Fernando, Mark Johnson, and Stephen Gould.
\newblock Spice: Semantic propositional image caption evaluation.
\newblock In {\em ECCV}, 2016.

\bibitem{anderson2018}
Peter Anderson, Xiaodong He, Chris Buehler, Damien Teney, Mark Johnson, Stephen
  Gould, and Lei Zhang.
\newblock Bottom-up and top-down attention for image captioning and visual
  question answering.
\newblock In {\em CVPR}, 2018.

\bibitem{andreas2016reasoning}
Jacob Andreas and Dan Klein.
\newblock Reasoning about pragmatics with neural listeners and speakers.
\newblock In {\em EMNLP}, 2016.

\bibitem{bengio2015scheduled}
Samy Bengio, Oriol Vinyals, Navdeep Jaitly, and Noam Shazeer.
\newblock Scheduled sampling for sequence prediction with recurrent neural
  networks.
\newblock In {\em NeurIPS}, 2015.

\bibitem{chen2018factual}
Tianlang Chen, Zhongping Zhang, Quanzeng You, Chen Fang, Zhaowen Wang, Hailin
  Jin, and Jiebo Luo.
\newblock ``factual''or``emotional'': Stylized image captioning with adaptive
  learning and attention.
\newblock In {\em ECCV}, 2018.

\bibitem{chen2015microsoft}
Xinlei Chen, Hao Fang, Tsung-Yi Lin, Ramakrishna Vedantam, Saurabh Gupta, Piotr
  Doll{\'a}r, and C~Lawrence Zitnick.
\newblock Microsoft coco captions: Data collection and evaluation server.
\newblock {\em arXiv preprint arXiv:1504.00325}, 2015.

\bibitem{cho-etal-2022-fine}
Jaemin Cho, Seunghyun Yoon, Ajinkya Kale, Franck Dernoncourt, Trung Bui, and
  Mohit Bansal.
\newblock Fine-grained image captioning with {CLIP} reward.
\newblock In {\em Findings of the Association for Computational Linguistics:
  NAACL 2022}, 2022.

\bibitem{choshen2019weaknesses}
Leshem Choshen, Lior Fox, Zohar Aizenbud, and Omri Abend.
\newblock On the weaknesses of reinforcement learning for neural machine
  translation.
\newblock In {\em ICLR}, 2020.

\bibitem{clark2019don}
Christopher Clark, Mark Yatskar, and Luke Zettlemoyer.
\newblock Don’t take the easy way out: Ensemble based methods for avoiding
  known dataset biases.
\newblock In {\em EMNLP-IJCNLP}, 2019.

\bibitem{cohn2018pragmatically}
Reuben Cohn-Gordon, Noah Goodman, and Christopher Potts.
\newblock Pragmatically informative image captioning with character-level
  inference.
\newblock In {\em {NAACL-HLT}}, 2018.

\bibitem{dai2017towards}
Bo Dai, Sanja Fidler, Raquel Urtasun, and Dahua Lin.
\newblock Towards diverse and natural image descriptions via a conditional gan.
\newblock In {\em ICCV}, 2017.

\bibitem{dai2017contrastive}
Bo Dai and Dahua Lin.
\newblock Contrastive learning for image captioning.
\newblock In {\em NeurIPS}, 2017.

\bibitem{demeter2020stolen}
David Demeter, Gregory Kimmel, and Doug Downey.
\newblock Stolen probability: A structural weakness of neural language models.
\newblock In {\em ACL}, 2020.

\bibitem{edunov2018classical}
Sergey Edunov, Myle Ott, Michael Auli, David Grangier, and Marc’Aurelio
  Ranzato.
\newblock Classical structured prediction losses for sequence to sequence
  learning.
\newblock In {\em NAACL-HLT}, 2018.

\bibitem{faghri2018vsepp}
Fartash Faghri, David~J Fleet, Jamie~Ryan Kiros, and Sanja Fidler.
\newblock Vse++: Improving visual-semantic embeddings with hard negatives.
\newblock In {\em BMVC}, 2018.

\bibitem{fisch2020}
Adam Fisch, Kenton Lee, Ming-Wei Chang, Jonathan~H Clark, and Regina Barzilay.
\newblock Capwap: Captioning with a purpose.
\newblock In {\em EMNLP}, 2020.

\bibitem{goodfellow2014}
Ian Goodfellow, Jean Pouget-Abadie, Mehdi Mirza, Bing Xu, David Warde-Farley,
  Sherjil Ozair, Aaron Courville, and Yoshua Bengio.
\newblock Generative adversarial nets.
\newblock In {\em NeurIPS}, 2014.

\bibitem{gu2020token}
Shuhao Gu, Jinchao Zhang, Fandong Meng, Yang Feng, Wanying Xie, Jie Zhou, and
  Dong Yu.
\newblock Token-level adaptive training for neural machine translation.
\newblock In {\em EMNLP}, 2020.

\bibitem{gurari2020}
Danna Gurari, Yinan Zhao, Meng Zhang, and Nilavra Bhattacharya.
\newblock Captioning images taken by people who are blind.
\newblock In {\em ECCV}, 2020.

\bibitem{he2019unlearn}
He He, Sheng Zha, and Haohan Wang.
\newblock Unlearn dataset bias in natural language inference by fitting the
  residual.
\newblock In {\em EMNLP-IJCNLP}, 2019.

\bibitem{hessel2021clipscore}
Jack Hessel, Ari Holtzman, Maxwell Forbes, Ronan~Le Bras, and Yejin Choi.
\newblock Clipscore: A reference-free evaluation metric for image captioning.
\newblock In {\em EMNLP}, 2021.

\bibitem{hinton2002training}
Geoffrey~E Hinton.
\newblock Training products of experts by minimizing contrastive divergence.
\newblock {\em Neural computation}, 14(8):1771--1800, 2002.

\bibitem{hochreiter1997}
Sepp Hochreiter and J{\"u}rgen Schmidhuber.
\newblock Long short-term memory.
\newblock {\em Neural computation}, 9(8):1735--1780, 1997.

\bibitem{holtzman2020curious}
Ari Holtzman, Jan Buys, Li Du, Maxwell Forbes, and Yejin Choi.
\newblock The curious case of neural text degeneration.
\newblock In {\em ICLR}, 2020.

\bibitem{jiang2019tiger}
Ming Jiang, Qiuyuan Huang, Lei Zhang, Xin Wang, Pengchuan Zhang, Zhe Gan, Jana
  Diesner, and Jianfeng Gao.
\newblock Tiger: Text-to-image grounding for image caption evaluation.
\newblock In {\em EMNLP-IJCNLP}, 2019.

\bibitem{jiang2019improving}
Shaojie Jiang, Pengjie Ren, Christof Monz, and Maarten de Rijke.
\newblock Improving neural response diversity with frequency-aware
  cross-entropy loss.
\newblock In {\em WWW}, 2019.

\bibitem{kang2019decoupling}
Bingyi Kang, Saining Xie, Marcus Rohrbach, Zhicheng Yan, Albert Gordo, Jiashi
  Feng, and Yannis Kalantidis.
\newblock Decoupling representation and classifier for long-tailed recognition.
\newblock In {\em ICLR}, 2020.

\bibitem{karpathy2015}
Andrej Karpathy and Li Fei-Fei.
\newblock Deep visual-semantic alignments for generating image descriptions.
\newblock In {\em CVPR}, 2015.

\bibitem{kasai2022transparent}
Jungo Kasai, Keisuke Sakaguchi, Lavinia Dunagan, Jacob Morrison, Ronan Le~Bras,
  Yejin Choi, and Noah~A. Smith.
\newblock Transparent human evaluation for image captioning.
\newblock In {\em NAACL-HLT}, 2022.

\bibitem{kiegeland2021revisiting}
Samuel Kiegeland and Julia Kreutzer.
\newblock Revisiting the weaknesses of reinforcement learning for neural
  machine translation.
\newblock In {\em NAACL-HLT}, 2021.

\bibitem{kim2020dense}
Hyounghun Kim, Zineng Tang, and Mohit Bansal.
\newblock Dense-caption matching and frame-selection gating for temporal
  localization in videoqa.
\newblock In {\em ACL}, 2020.

\bibitem{krishna2017}
Ranjay Krishna, Yuke Zhu, Oliver Groth, Justin Johnson, Kenji Hata, Joshua
  Kravitz, Stephanie Chen, Yannis Kalantidis, Li-Jia Li, David~A Shamma,
  Michael Bernstein, and Li Fei-Fei.
\newblock Visual genome: Connecting language and vision using crowdsourced
  dense image annotations.
\newblock {\em IJCV}, 123(1):32--73, 2017.

\bibitem{lee-etal-2020-vilbertscore}
Hwanhee Lee, Seunghyun Yoon, Franck Dernoncourt, Doo~Soon Kim, Trung Bui, and
  Kyomin Jung.
\newblock {V}i{LBERTS}core: Evaluating image caption using vision-and-language
  {BERT}.
\newblock In {\em The First Workshop on Evaluation and Comparison of NLP
  Systems}, 2020.

\bibitem{li-etal-2016-deep}
Jiwei Li, Will Monroe, Alan Ritter, Dan Jurafsky, Michel Galley, and Jianfeng
  Gao.
\newblock Deep reinforcement learning for dialogue generation.
\newblock In {\em EMNLP}, 2016.

\bibitem{li2020dice}
Xiaoya Li, Xiaofei Sun, Yuxian Meng, Junjun Liang, Fei Wu, and Jiwei Li.
\newblock Dice loss for data-imbalanced nlp tasks.
\newblock In {\em ACL}, 2020.

\bibitem{li2020oscar}
Xiujun Li, Xi Yin, Chunyuan Li, Pengchuan Zhang, Xiaowei Hu, Lei Zhang, Lijuan
  Wang, Houdong Hu, Li Dong, Furu Wei, et~al.
\newblock Oscar: Object-semantics aligned pre-training for vision-language
  tasks.
\newblock In {\em ECCV}, 2020.

\bibitem{lin2017focal}
Tsung-Yi Lin, Priya Goyal, Ross Girshick, Kaiming He, and Piotr Doll{\'a}r.
\newblock Focal loss for dense object detection.
\newblock In {\em ICCV}, 2017.

\bibitem{lin2014mscoco}
Tsung-Yi Lin, Michael Maire, Serge Belongie, James Hays, Pietro Perona, Deva
  Ramanan, Piotr Doll{\'a}r, and C~Lawrence Zitnick.
\newblock Microsoft coco: Common objects in context.
\newblock In {\em ECCV}, 2014.

\bibitem{liu2019generating}
Lixin Liu, Jiajun Tang, Xiaojun Wan, and Zongming Guo.
\newblock Generating diverse and descriptive image captions using visual
  paraphrases.
\newblock In {\em ICCV}, 2019.

\bibitem{liu2018show}
Xihui Liu, Hongsheng Li, Jing Shao, Dapeng Chen, and Xiaogang Wang.
\newblock Show, tell and discriminate: Image captioning by self-retrieval with
  partially labeled data.
\newblock In {\em ECCV}, 2018.

\bibitem{luo2018discriminability}
Ruotian Luo, Brian Price, Scott Cohen, and Gregory Shakhnarovich.
\newblock Discriminability objective for training descriptive captions.
\newblock In {\em CVPR}, 2018.

\bibitem{luo2020analysis}
Ruotian Luo and Gregory Shakhnarovich.
\newblock Analysis of diversity-accuracy tradeoff in image captioning.
\newblock {\em arXiv preprint arXiv:2002.11848}, 2020.

\bibitem{mao2022rethinking}
Yangjun Mao, Long Chen, Zhihong Jiang, Dong Zhang, Zhimeng Zhang, Jian Shao,
  and Jun Xiao.
\newblock Rethinking the reference-based distinctive image captioning.
\newblock In {\em ACM MM}, 2022.

\bibitem{menon2020long}
Aditya~Krishna Menon, Sadeep Jayasumana, Ankit~Singh Rawat, Himanshu Jain,
  Andreas Veit, and Sanjiv Kumar.
\newblock Long-tail learning via logit adjustment.
\newblock In {\em ICLR}, 2020.

\bibitem{monroe2017colors}
Will Monroe, Robert~XD Hawkins, Noah~D Goodman, and Christopher Potts.
\newblock Colors in context: A pragmatic neural model for grounded language
  understanding.
\newblock {\em TACL}, 5:325--338, 2017.

\bibitem{nguyen2018improving}
Toan~Q Nguyen and David Chiang.
\newblock Improving lexical choice in neural machine translation.
\newblock In {\em NAACL-HLT}, 2018.

\bibitem{pasunuru-bansal-2018-multi}
Ramakanth Pasunuru and Mohit Bansal.
\newblock Multi-reward reinforced summarization with saliency and entailment.
\newblock In {\em NAACL-HLT}. ACL, 2018.

\bibitem{ranzato2015}
Marc'Aurelio Ranzato, Sumit Chopra, Michael Auli, and Wojciech Zaremba.
\newblock Sequence level training with recurrent neural networks.
\newblock In {\em ICLR}, 2015.

\bibitem{raunak-etal-2020-long}
Vikas Raunak, Siddharth Dalmia, Vivek Gupta, and Florian Metze.
\newblock On long-tailed phenomena in neural machine translation.
\newblock In {\em Findings of the Association for Computational Linguistics:
  EMNLP 2020}, 2020.

\bibitem{rennie2017}
Steven~J Rennie, Etienne Marcheret, Youssef Mroueh, Jerret Ross, and Vaibhava
  Goel.
\newblock Self-critical sequence training for image captioning.
\newblock In {\em CVPR}, 2017.

\bibitem{sadovnik2012image}
Amir Sadovnik, Yi-I Chiu, Noah Snavely, Shimon Edelman, and Tsuhan Chen.
\newblock Image description with a goal: Building efficient discriminating
  expressions for images.
\newblock In {\em CVPR}, 2012.

\bibitem{sharma2018}
Piyush Sharma, Nan Ding, Sebastian Goodman, and Radu Soricut.
\newblock Conceptual captions: A cleaned, hypernymed, image alt-text dataset
  for automatic image captioning.
\newblock In {\em ACL}, 2018.

\bibitem{shi2021partial}
Jiahe Shi, Yali Li, and Shengjin Wang.
\newblock Partial off-policy learning: Balance accuracy and diversity for
  human-oriented image captioning.
\newblock In {\em ICCV}, 2021.

\bibitem{shi-etal-2021-enhancing}
Zhan Shi, Hui Liu, and Xiaodan Zhu.
\newblock Enhancing descriptive image captioning with natural language
  inference.
\newblock In {\em ACL}, 2021.

\bibitem{stefanini2021show}
Matteo Stefanini, Marcella Cornia, Lorenzo Baraldi, Silvia Cascianelli,
  Giuseppe Fiameni, and Rita Cucchiara.
\newblock From show to tell: A survey on image captioning.
\newblock {\em arXiv preprint arXiv:2107.06912}, 2021.

\bibitem{tang2020long}
Kaihua Tang, Jianqiang Huang, and Hanwang Zhang.
\newblock Long-tailed classification by keeping the good and removing the bad
  momentum causal effect.
\newblock In {\em NeurIPS}, 2020.

\bibitem{vaswani2017attention}
Ashish Vaswani, Noam Shazeer, Niki Parmar, Jakob Uszkoreit, Llion Jones,
  Aidan~N Gomez, {\L}ukasz Kaiser, and Illia Polosukhin.
\newblock Attention is all you need.
\newblock In {\em NeurIPS}, 2017.

\bibitem{vedantam2017context}
Ramakrishna Vedantam, Samy Bengio, Kevin Murphy, Devi Parikh, and Gal Chechik.
\newblock Context-aware captions from context-agnostic supervision.
\newblock In {\em CVPR}, 2017.

\bibitem{vedantam2015}
Ramakrishna Vedantam, C Lawrence~Zitnick, and Devi Parikh.
\newblock Cider: Consensus-based image description evaluation.
\newblock In {\em CVPR}, 2015.

\bibitem{vered2019joint}
Gilad Vered, Gal Oren, Yuval Atzmon, and Gal Chechik.
\newblock Joint optimization for cooperative image captioning.
\newblock In {\em ICCV}, 2019.

\bibitem{vinyals2015}
Oriol Vinyals, Alexander Toshev, Samy Bengio, and Dumitru Erhan.
\newblock Show and tell: A neural image caption generator.
\newblock In {\em CVPR}, 2015.

\bibitem{wang2020compare}
Jiuniu Wang, Wenjia Xu, Qingzhong Wang, and Antoni~B Chan.
\newblock Compare and reweight: Distinctive image captioning using similar
  images sets.
\newblock In {\em ECCV}, 2020.

\bibitem{wang2021group}
Jiuniu Wang, Wenjia Xu, Qingzhong Wang, and Antoni~B Chan.
\newblock Group-based distinctive image captioning with memory attention.
\newblock In {\em ACM MM}, 2021.

\bibitem{wang2019describing}
Qingzhong Wang and Antoni~B Chan.
\newblock Describing like humans: on diversity in image captioning.
\newblock In {\em CVPR}, 2019.

\bibitem{wang2020long}
Xudong Wang, Long Lian, Zhongqi Miao, Ziwei Liu, and Stella Yu.
\newblock Long-tailed recognition by routing diverse distribution-aware
  experts.
\newblock In {\em ICLR}, 2020.

\bibitem{wang2020towards}
Zeyu Wang, Berthy Feng, Karthik Narasimhan, and Olga Russakovsky.
\newblock Towards unique and informative captioning of images.
\newblock In {\em ECCV}, 2020.

\bibitem{wang2016diverse}
Zhuhao Wang, Fei Wu, Weiming Lu, Jun Xiao, Xi Li, Zitong Zhang, and Yueting
  Zhuang.
\newblock Diverse image captioning via grouptalk.
\newblock In {\em IJCAI}, 2016.

\bibitem{white2021open}
Julia White, Gabriel Poesia, Robert Hawkins, Dorsa Sadigh, and Noah Goodman.
\newblock Open-domain clarification question generation without question
  examples.
\newblock In {\em EMNLP}, 2021.

\bibitem{williams1992}
Ronald~J Williams.
\newblock Simple statistical gradient-following algorithms for connectionist
  reinforcement learning.
\newblock {\em Machine learning}, 8(3):229--256, 1992.

\bibitem{wu2020fine}
Jie Wu, Tianshui Chen, Hefeng Wu, Zhi Yang, Guangchun Luo, and Liang Lin.
\newblock Fine-grained image captioning with global-local discriminative
  objective.
\newblock {\em IEEE Transactions on Multimedia}, 23:2413--2427, 2021.

\bibitem{wu-etal-2018-study}
Lijun Wu, Fei Tian, Tao Qin, Jianhuang Lai, and Tie-Yan Liu.
\newblock A study of reinforcement learning for neural machine translation.
\newblock In {\em EMNLP}, 2018.

\bibitem{wu2020importance}
Qingyang Wu, Lei Li, Hao Zhou, Ying Zeng, and Zhou Yu.
\newblock Importance-aware learning for neural headline editing.
\newblock In {\em AAAI}, 2020.

\bibitem{xu2015}
Kelvin Xu, Jimmy Ba, Ryan Kiros, Kyunghyun Cho, Aaron Courville, Ruslan
  Salakhudinov, Rich Zemel, and Yoshua Bengio.
\newblock Show, attend and tell: Neural image caption generation with visual
  attention.
\newblock In {\em ICML}, 2015.

\bibitem{yi2020improving}
Yanzhi Yi, Hangyu Deng, and Jinglu Hu.
\newblock Improving image captioning evaluation by considering inter references
  variance.
\newblock In {\em ACL}, 2020.

\bibitem{young2014image}
Peter Young, Alice Lai, Micah Hodosh, and Julia Hockenmaier.
\newblock From image descriptions to visual denotations: New similarity metrics
  for semantic inference over event descriptions.
\newblock {\em TACL}, 2:67--78, 2014.

\bibitem{zhang2021vinvl}
Pengchuan Zhang, Xiujun Li, Xiaowei Hu, Jianwei Yang, Lei Zhang, Lijuan Wang,
  Yejin Choi, and Jianfeng Gao.
\newblock Vinvl: Revisiting visual representations in vision-language models.
\newblock In {\em CVPR}, 2021.

\bibitem{zhang2020bertscore}
Tianyi Zhang, Varsha Kishore, Felix Wu, Kilian~Q Weinberger, and Yoav Artzi.
\newblock Bertscore: Evaluating text generation with bert.
\newblock In {\em ICLR}, 2020.

\bibitem{zhang2021deep}
Yifan Zhang, Bingyi Kang, Bryan Hooi, Shuicheng Yan, and Jiashi Feng.
\newblock Deep long-tailed learning: A survey.
\newblock {\em arXiv preprint arXiv:2110.04596}, 2021.

\bibitem{zhang2021show}
Zhongping Zhang, Yiwen Gu, and Bryan~A Plummer.
\newblock Show and write: Entity-aware news generation with image information.
\newblock {\em arXiv preprint arXiv:2112.05917}, 2021.

\end{thebibliography}


\begin{thebibliography}{10}\itemsep=-1pt

\bibitem{anderson2018}
Peter Anderson, Xiaodong He, Chris Buehler, Damien Teney, Mark Johnson, Stephen
  Gould, and Lei Zhang.
\newblock Bottom-up and top-down attention for image captioning and visual
  question answering.
\newblock In {\em CVPR}, 2018.

\bibitem{basu2020mirostat}
Sourya Basu, Govardana~Sachitanandam Ramachandran, Nitish~Shirish Keskar, and
  Lav~R Varshney.
\newblock Mirostat: A neural text decoding algorithm that directly controls
  perplexity.
\newblock In {\em ICLR}, 2021.

\bibitem{cho-etal-2022-fine}
Jaemin Cho, Seunghyun Yoon, Ajinkya Kale, Franck Dernoncourt, Trung Bui, and
  Mohit Bansal.
\newblock Fine-grained image captioning with {CLIP} reward.
\newblock In {\em Findings of the Association for Computational Linguistics:
  NAACL 2022}, 2022.

\bibitem{fellbaum1998wordnet}
Christiane Fellbaum.
\newblock {\em {WordNet: An Electronic Lexical Database}}.
\newblock The MIT Press, 1998.

\bibitem{gu2020token}
Shuhao Gu, Jinchao Zhang, Fandong Meng, Yang Feng, Wanying Xie, Jie Zhou, and
  Dong Yu.
\newblock Token-level adaptive training for neural machine translation.
\newblock In {\em EMNLP}, 2020.

\bibitem{hendricks2018women}
Lisa~Anne Hendricks, Kaylee Burns, Kate Saenko, Trevor Darrell, and Anna
  Rohrbach.
\newblock Women also snowboard: Overcoming bias in captioning models.
\newblock In {\em ECCV}, 2018.

\bibitem{holtzman2020curious}
Ari Holtzman, Jan Buys, Li Du, Maxwell Forbes, and Yejin Choi.
\newblock The curious case of neural text degeneration.
\newblock In {\em ICLR}, 2020.

\bibitem{jiang2019improving}
Shaojie Jiang, Pengjie Ren, Christof Monz, and Maarten de Rijke.
\newblock Improving neural response diversity with frequency-aware
  cross-entropy loss.
\newblock In {\em WWW}, 2019.

\bibitem{kang2019decoupling}
Bingyi Kang, Saining Xie, Marcus Rohrbach, Zhicheng Yan, Albert Gordo, Jiashi
  Feng, and Yannis Kalantidis.
\newblock Decoupling representation and classifier for long-tailed recognition.
\newblock In {\em ICLR}, 2020.

\bibitem{li2020oscar}
Xiujun Li, Xi Yin, Chunyuan Li, Pengchuan Zhang, Xiaowei Hu, Lei Zhang, Lijuan
  Wang, Houdong Hu, Li Dong, Furu Wei, et~al.
\newblock Oscar: Object-semantics aligned pre-training for vision-language
  tasks.
\newblock In {\em ECCV}, 2020.

\bibitem{lin2017focal}
Tsung-Yi Lin, Priya Goyal, Ross Girshick, Kaiming He, and Piotr Doll{\'a}r.
\newblock Focal loss for dense object detection.
\newblock In {\em ICCV}, 2017.

\bibitem{liu2019generating}
Lixin Liu, Jiajun Tang, Xiaojun Wan, and Zongming Guo.
\newblock Generating diverse and descriptive image captions using visual
  paraphrases.
\newblock In {\em ICCV}, 2019.

\bibitem{meister2022typical}
Clara Meister, Tiago Pimentel, Gian Wiher, and Ryan Cotterell.
\newblock Typical decoding for natural language generation.
\newblock {\em arXiv preprint arXiv:2202.00666}, 2022.

\bibitem{nguyen2018improving}
Toan~Q Nguyen and David Chiang.
\newblock Improving lexical choice in neural machine translation.
\newblock In {\em NAACL-HLT}, 2018.

\bibitem{nguyen2022grit}
Van-Quang Nguyen, Masanori Suganuma, and Takayuki Okatani.
\newblock Grit: Faster and better image captioning transformer using dual
  visual features.
\newblock In {\em ECCV}, 2022.

\bibitem{pmlr-v139-radford21a}
Alec Radford, Jong~Wook Kim, Chris Hallacy, Aditya Ramesh, Gabriel Goh,
  Sandhini Agarwal, Girish Sastry, Amanda Askell, Pamela Mishkin, Jack Clark,
  Gretchen Krueger, and Ilya Sutskever.
\newblock Learning transferable visual models from natural language
  supervision.
\newblock In {\em ICML}, 2021.

\bibitem{raunak-etal-2020-long}
Vikas Raunak, Siddharth Dalmia, Vivek Gupta, and Florian Metze.
\newblock On long-tailed phenomena in neural machine translation.
\newblock In {\em Findings of the Association for Computational Linguistics:
  EMNLP 2020}, 2020.

\bibitem{rennie2017}
Steven~J Rennie, Etienne Marcheret, Youssef Mroueh, Jerret Ross, and Vaibhava
  Goel.
\newblock Self-critical sequence training for image captioning.
\newblock In {\em CVPR}, 2017.

\bibitem{sadovnik2012image}
Amir Sadovnik, Yi-I Chiu, Noah Snavely, Shimon Edelman, and Tsuhan Chen.
\newblock Image description with a goal: Building efficient discriminating
  expressions for images.
\newblock In {\em CVPR}, 2012.

\bibitem{stefanini2021show}
Matteo Stefanini, Marcella Cornia, Lorenzo Baraldi, Silvia Cascianelli,
  Giuseppe Fiameni, and Rita Cucchiara.
\newblock From show to tell: A survey on image captioning.
\newblock {\em arXiv preprint arXiv:2107.06912}, 2021.

\bibitem{wang2022ofa}
Peng Wang, An Yang, Rui Men, Junyang Lin, Shuai Bai, Zhikang Li, Jianxin Ma,
  Chang Zhou, Jingren Zhou, and Hongxia Yang.
\newblock Ofa: Unifying architectures, tasks, and modalities through a simple
  sequence-to-sequence learning framework.
\newblock In {\em ICML}, 2022.

\bibitem{wu2020importance}
Qingyang Wu, Lei Li, Hao Zhou, Ying Zeng, and Zhou Yu.
\newblock Importance-aware learning for neural headline editing.
\newblock In {\em AAAI}, 2020.

\bibitem{zhang2021vinvl}
Pengchuan Zhang, Xiujun Li, Xiaowei Hu, Jianwei Yang, Lei Zhang, Lijuan Wang,
  Yejin Choi, and Jianfeng Gao.
\newblock Vinvl: Revisiting visual representations in vision-language models.
\newblock In {\em CVPR}, 2021.

\bibitem{zhao2021captionbias}
Dora Zhao, Angelina Wang, and Olga Russakovsky.
\newblock Understanding and evaluating racial biases in image captioning.
\newblock In {\em ICCV}, 2021.

\bibitem{zhao-etal-2017-men}
Jieyu Zhao, Tianlu Wang, Mark Yatskar, Vicente Ordonez, and Kai-Wei Chang.
\newblock Men also like shopping: Reducing gender bias amplification using
  corpus-level constraints.
\newblock In {\em EMNLP}, 2017.

\end{thebibliography}
}

\end{document}


\title{Switching to Discriminative Image Captioning\\by Relieving a Bottleneck of Reinforcement Learning\\-- Supplementary Material}

\author{
Ukyo Honda$^{1,2}$\qquad Taro Watanabe$^3$\qquad Yuji Matsumoto$^2$\\
$^1$CyberAgent, Inc.\qquad $^2$RIKEN\qquad $^3$Nara Institute of Science and Technology\qquad\\
\small{
\texttt{honda\_ukyo@cyberagent.co.jp}\qquad \texttt{taro@is.naist.jp}\qquad \texttt{yuji.matsumoto@riken.jp}
}
}

\maketitle
\thispagestyle{empty}


\begin{figure}[t]
    \centering
    \includegraphics[width=0.9\columnwidth,keepaspectratio]{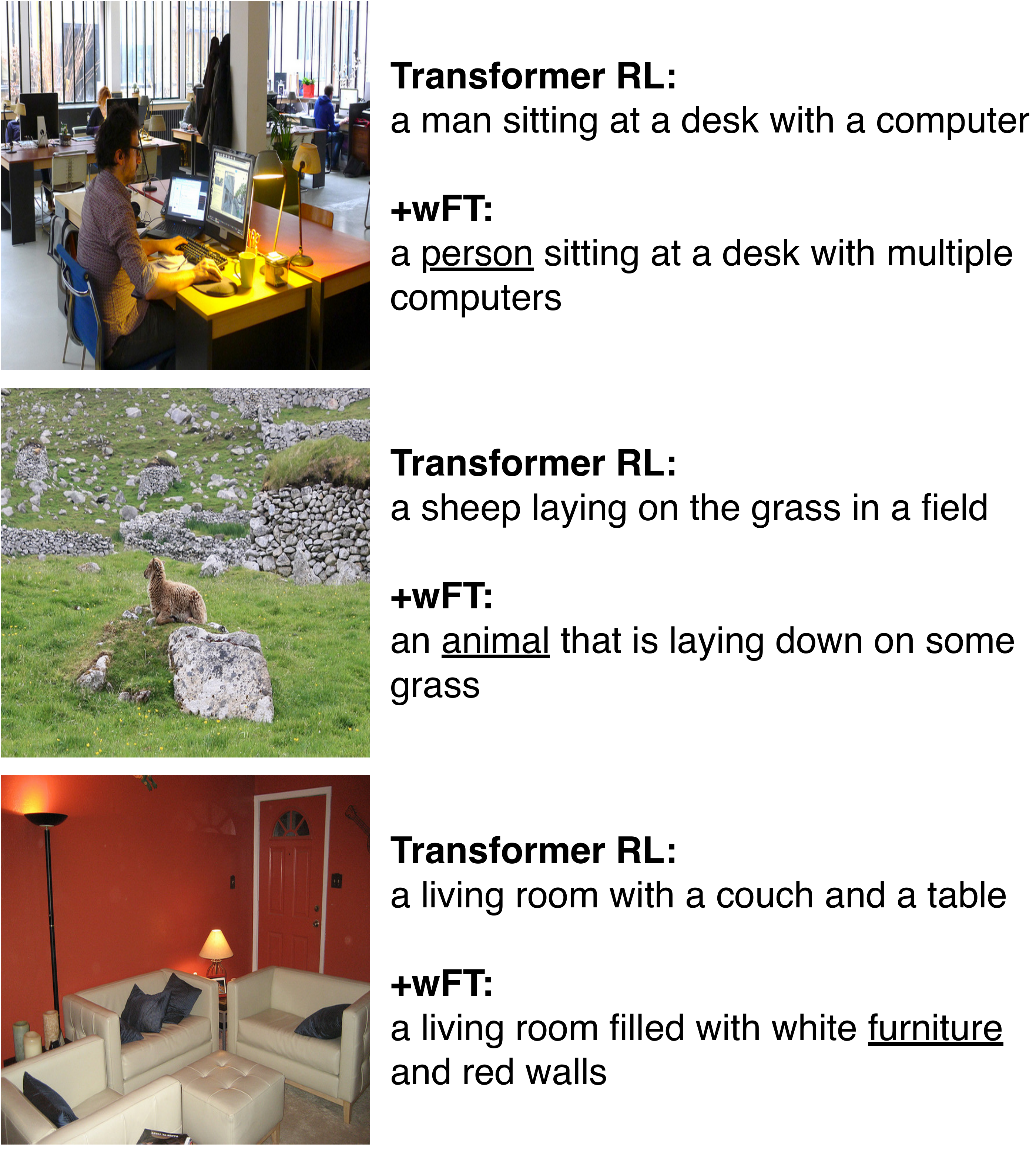}
    \caption{
    Examples of the limitation of our methods.
    All the examples are from the MS~COCO validation set.
    The underlined words are relatively low-frequency hypernyms.
    }
    \label{fig:limitation}
\end{figure}

\section{Limitations and Ethical Considerations}
Our experiments were limited to the MS~COCO dataset, which is the standard dataset for image captioning.
The images belong to the general domain (real images of common objects), and the captions are in English only.
To compensate for the limitation, we have demonstrated the effectiveness of our methods with the multiple baseline models.

Our current methods have a limitation in that they cannot select discriminative ones among low-frequency words.
Although discriminative in general, low-frequency words do not always describe more specific information than others.
Figure~\ref{fig:limitation} shows the examples.
Our model output relatively low-frequency hypernyms such as \emph{person}, \emph{animal}, and \emph{furniture} instead of the more frequent but more specific hyponyms: \emph{man}, \emph{sheep}, and \emph{couch}.
Utilizing thesauruses like WordNet~\cite{fellbaum1998wordnet} will be a promising approach to reduce those relatively low-frequency hypernyms from outputs.

The dataset contains social biases, and captioning models have the risk of amplifying those biases~\cite{zhao2021captionbias,zhao-etal-2017-men,hendricks2018women}.
Our methods are also not free from the risk, as they are not designed to reduce those social biases from existing models.

\begin{figure}[t]
    \centering
    \includegraphics[width=1.0\columnwidth,keepaspectratio]{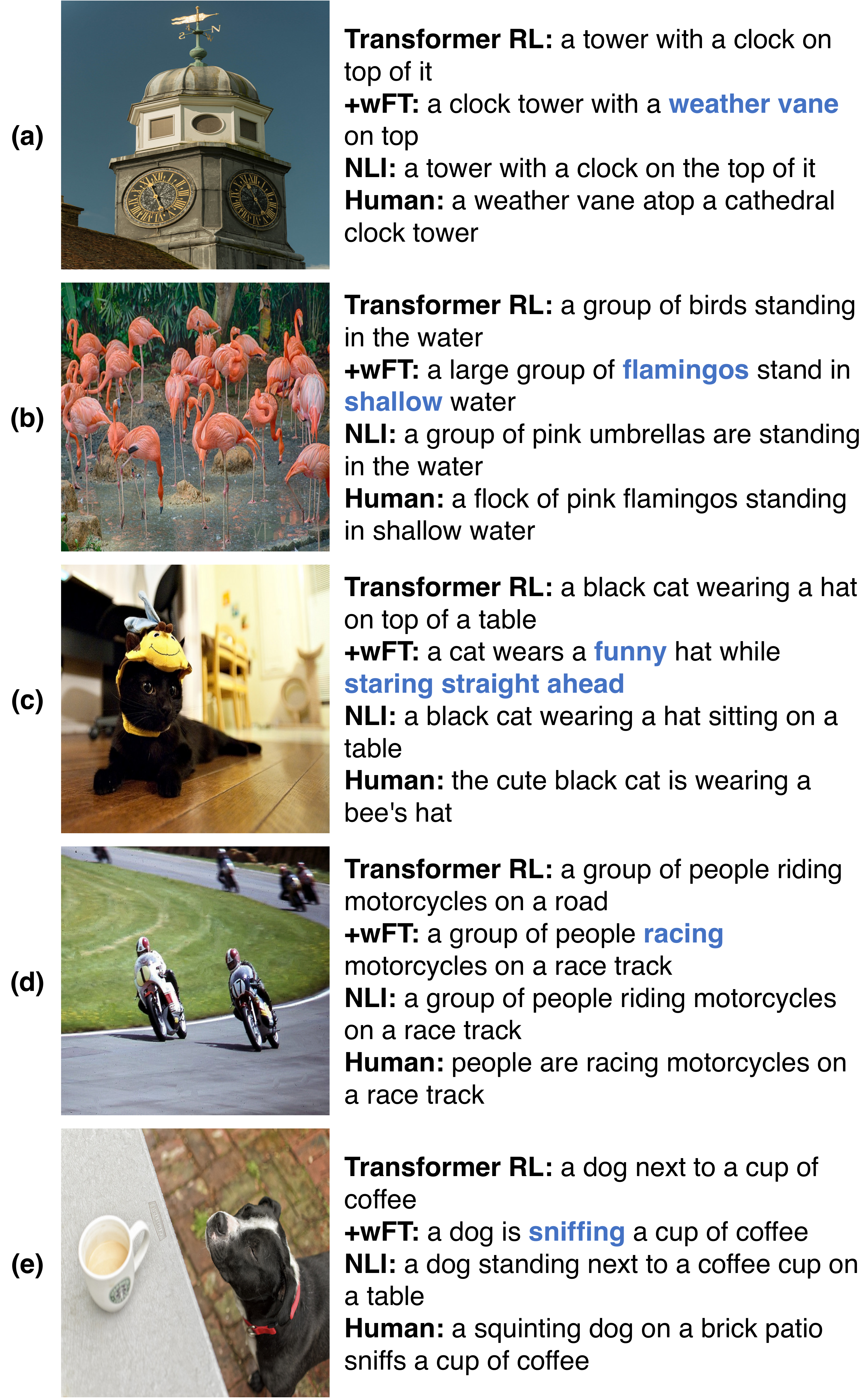}
    \caption{
    Caption examples in the MS~COCO validation set.
    The blue words are those that have never appeared in the output captions of the baseline model (Transformer RL).
    \emph{Human} shows a ground-truth caption of each image.
    }
    \label{fig:qualitative}
\end{figure}

\section{Further Output Examples}
\label{appendix:qualitative}
Figure~\ref{fig:qualitative} shows caption examples in the MS~COCO validation set.
The blue words are those that have never appeared in the output captions of the baseline model.
We observe that these blue words express various types of characteristic information of the images.
Here, \emph{weather vane} and \emph{flamingos} are characteristic objects of the images (a) and (b); \emph{shallow}, \emph{funny}, and \emph{staring straight ahead} are characteristic attributes of the images (b) and (c); and \emph{racing} and \emph{sniffing} are characteristic relations in the images (d) and (e).
These examples further support our hypothesis that the limited vocabulary of RL models hinders discriminativeness.

\begin{figure}[t]
    \centering
    \includegraphics[width=1.0\columnwidth,keepaspectratio]{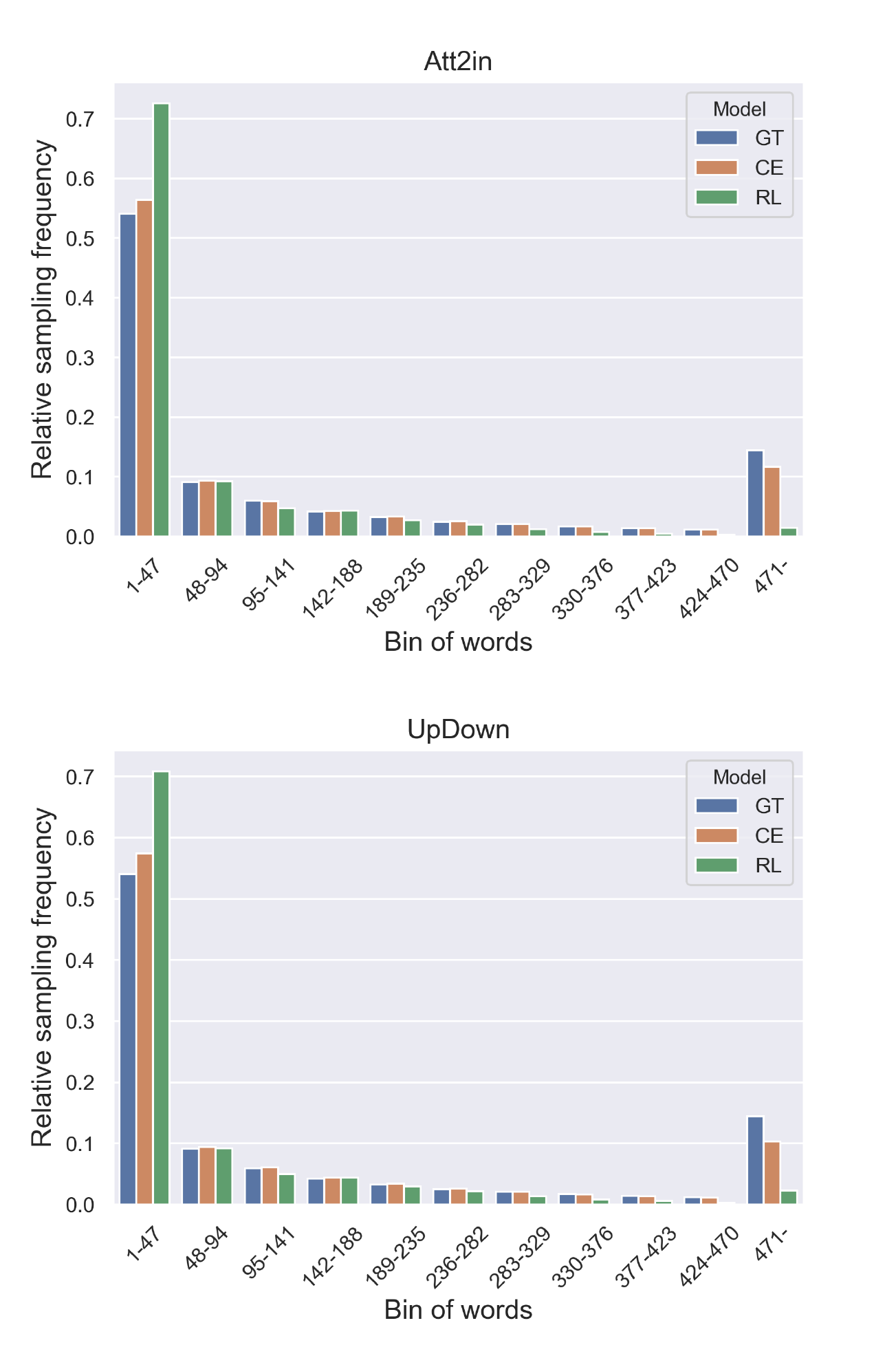}
    \caption{
    Relative frequency of the words in the sequences sampled for the training images.
    Five sequences were sampled for each image.
    The words (9,486 unique words excluding an out-of-vocabulary token $\langle \mathrm{unk} \rangle$) are sorted by their frequency in ground-truth captions and divided into 200 bins.
    We show the first 10 bins and the sum of the rest.
    GT is the ground-truth caption of the training images, CE is the output of a captioning model trained with the CE loss, and RL is the output of a captioning model trained with RL.
    }
    \label{fig:peakiness_appendix}
\end{figure}

\section{Peaky Distributions in Other Models}
\label{appendix:peakiness}
Figure~\ref{fig:peakiness_appendix} shows the results of the relative frequency of the words sampled for the training images by the LSTM-based models: Att2in~\cite{rennie2017} and UpDown~\cite{anderson2018}.
Similar to the Transformer model, the sequences sampled with the LSTM-based RL models are clearly limited to high-frequency words, forming the peaky distributions.

\begin{table}[t]
\centering
\scalebox{0.80}{
\begin{tabular}{lrrrr}  
\toprule
 & Epoch & Batch & Hour/Epoch & Total Hour \\
\midrule
\textbf{Att2in RL} & \textcolor[gray]{0.5}{20} & \textcolor[gray]{0.5}{10} & \textcolor[gray]{0.5}{0.68} & \textcolor[gray]{0.5}{13.54} \\
\rowcolor[rgb]{0.9, 0.9, 0.9}
\ \ + sFT & 1 & 10 & \textbf{0.08} & \textbf{0.08} \\
\rowcolor[rgb]{0.9, 0.9, 0.9}
\ \ + wFT & 1 & 10 & 0.12 & 0.12  \\
CIDErBtw & 50 & 10 & 0.70 & 35.11 \\
NLI & 50 & 16 & 0.87 & 43.55 \\
Joint CE & 20 & 10 & 1.15 & 22.97 \\
\midrule
\textbf{UpDown RL} & \textcolor[gray]{0.5}{20} & \textcolor[gray]{0.5}{10} & \textcolor[gray]{0.5}{0.71} & \textcolor[gray]{0.5}{14.16} \\
\rowcolor[rgb]{0.9, 0.9, 0.9}
\ \ + sFT & 1 & 10 & \textbf{0.09} & \textbf{0.09} \\
\rowcolor[rgb]{0.9, 0.9, 0.9}
\ \ + wFT & 1 & 10 & 0.14 & 0.14 \\
CIDErBtw & 50 & 10 & 0.76 & 38.09 \\
NLI & 50 & 16 & 0.87 & 43.74 \\
Joint CE & 20 & 10 & 1.08 & 21.67 \\
\midrule
\textbf{Transformer RL} & \textcolor[gray]{0.5}{25} & \textcolor[gray]{0.5}{10} & \textcolor[gray]{0.5}{3.23} & \textcolor[gray]{0.5}{80.66} \\
\rowcolor[rgb]{0.9, 0.9, 0.9}
\ \ + sFT & 1 & 10 & \textbf{0.11} & \textbf{0.11} \\
\rowcolor[rgb]{0.9, 0.9, 0.9}
\ \ + wFT & 1 & 10 & 0.18 & 0.18 \\
CIDErBtw & 25 & 10 & 3.27 & 81.76 \\
NLI & 25 & 16 & 2.74 & 68.54 \\
Joint CE & 25 & 10 & 4.06 & 101.43 \\
\bottomrule
\end{tabular}
}
\caption{
Time to train discriminativeness-aware captioning models.
Note that we excluded the time for initialization before RL because there is not much difference among the methods.
Results for the baseline RL models are shown in \textcolor[gray]{0.5}{gray text} because we did not train these models but used publicly-available pre-trained models.
}
\label{tab:time}
\end{table}

\section{Libraries for Evaluation}
\label{appendix:artifacts}
We used the following libraries for evaluation with all the hyperparameters set to the default values.

\noindent \textbf{CIDEr, SPICE, CLIPS, and RefCLIPS} \url{https://github.com/jmhessel/pycocoevalcap}

\noindent \textbf{BERTS+} \url{https://github.com/ck0123/improved-bertscore-for-image-captioning-evaluation}

\noindent \textbf{TIGEr} \url{https://github.com/SeleenaJM/CapEval}

\noindent \textbf{R@K} \url{https://github.com/fartashf/vsepp}; following \cite{liu2019generating}, we used a publicly available model, \texttt{coco\_vse++\_resnet\_restval\_finetune}.

\section{Best Hyperparameters}
\label{appendix:hyper}
We searched for the best hyperparameters for the learning rate from $\{1\textrm{e-}3,1\textrm{e-}4,1\textrm{e-}5,1\textrm{e-}6\}$, and the inverse-temperature hyperparameter $\beta'$ of Eq.~(\textcolor{red}{7}) from $\{0.1, 1\}$.
The best learning rate was 1e-5 for Transformer models and 1e-4 for the other models (Att2in and UpDown).
The best $\beta'$ was 0.1 for wFT with $p_{\theta}$ decoding and 1 for wFT with BP decoding.
Note that sFT does not use $\beta'$.

The best learning rate was the same in CE-based models (Joint CE and Only CE): 1e-5 for Transformer and 1e-4 for the others.
The best $\lambda \in \{0.2, 0.5, 0.8\}$ for Joint CE was 0.8 for Transformer and 0.2 for the others.

\section{The Number of Parameters}
\label{appendix:parameter size}
The exact number of parameters was 14,451,985 for Att2in, 52,125,025 for UpDown, and 57,474,832 for Transformer.
Note that the parameters $\theta'$ are not included because they are not trainable and fixed through the entire training and evaluation; rather, the actual trainable parameters are decreased to the classifier parameters in our models.
Visual Paraphrase has double decoders of Att2in; thus, it increases the number of trainable parameters and requires training of the specialized model from scratch.

\section{Comparison of Computational Cost}
\label{appendix:time}
Table~\ref{tab:time} shows the time to train discriminativeness-aware captioning models.
We used a single GPU of 16 GB memory for all training.
Clearly, our methods require far less time for training.
This is because our methods do not require retraining from scratch but only require a single-epoch fine-tuning to publicly-available pre-trained RL models.

\begin{figure}[t]
    \centering
    \includegraphics[width=1.0\columnwidth,keepaspectratio]{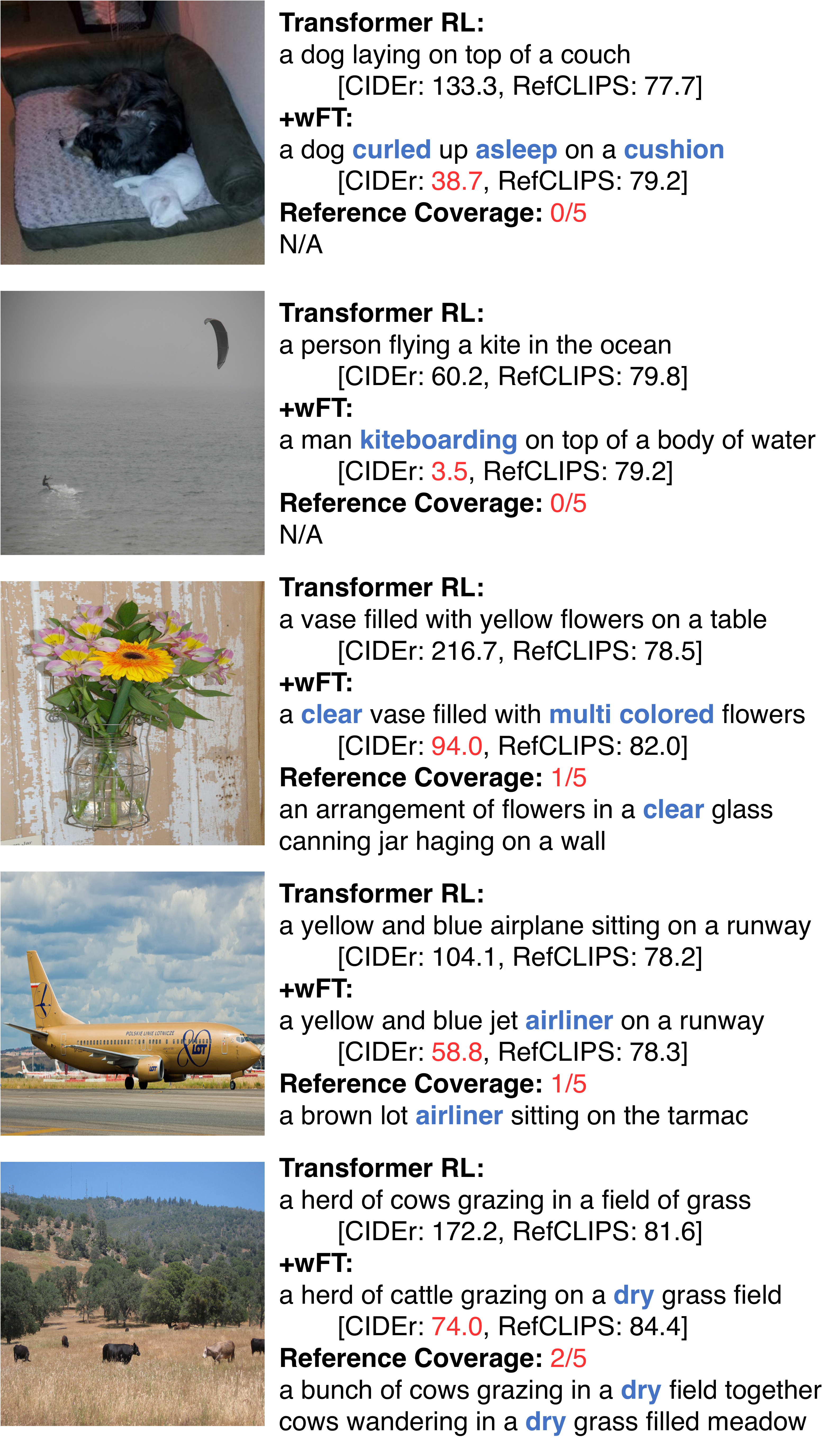}
    \caption{
    Underrated captions in the MS~COCO validation set.
    The blue words are those that have never appeared in the output captions of the baseline model (Transformer RL).
    \emph{Reference Coverage} shows the number of reference captions (out of five) that cover at least one of the blue words.
    }
    \label{fig:metric}
\end{figure}

\section{Qualitative Analysis of Underrated Captions}
\label{appendix:underrated qualitative}
Figure~\ref{fig:metric} shows caption examples, automatic evaluation scores, and reference captions.
Clearly, our wFT model correctly described all five images with diverse vocabulary.
However, the CIDEr scores for our captions were considerably lower than those for the baseline model captions.
The cause of this underrating is the small coverage of the reference captions: the reference captions rarely include the low-frequency words colored in blue due to their low frequency.
Conventional exact-matching metrics such as CIDEr cannot evaluate those correct-but-OOR words by the definition of exact-matching.
In contrast, RefCLIPS, the state-of-the-art soft-matching metric, can consider the information not covered by reference captions by incorporating image features.
Figure~\ref{fig:metric} shows that RefCLIPS evaluated the correct-but-OOR words more correctly and gave more plausible scores to our captions.
These examples further support our conclusion that the lower exact-matching scores of our models are caused by the nature of low-frequency words and the deficiency of exact-matching metrics, not by the degeneration of our models.

\begin{figure}[t]
    \centering
    \includegraphics[width=1.0\columnwidth,keepaspectratio]{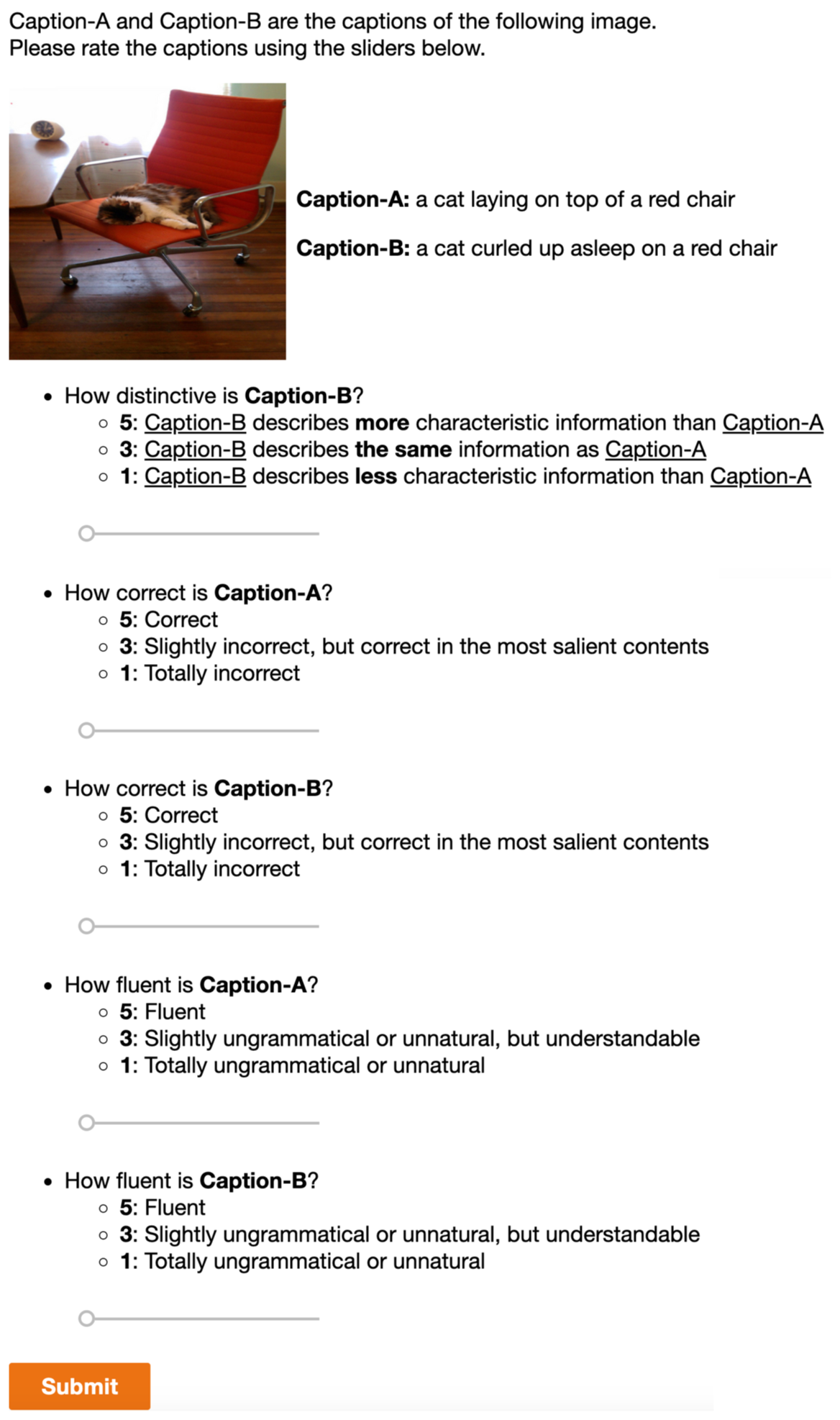}
    \caption{
    A screenshot of our AMT interface.
    }
    \label{fig:amt}
\end{figure}

\section{Details of Human Evaluation}
\label{appendix:protocol}
We show our AMT interface in Figure~\ref{fig:amt}.
Each image was evaluated with the five questions in the discrete 5-point scale.
We required workers to satisfy the following qualifications: being an AMT Master and living in the U.S.
Workers were notified that this experiment was intended to evaluate caption quality.
We paid \$0.1 for each image, and the median of the actual working time was 41 seconds per image.
The hourly reward was estimated as \$8.78, which is higher than the minimum wage in the U.S., \$7.25 per hour.

\section{Comparison with Other Long-Tail Classification Methods}
\label{appendix:long-tail}
We adapted the long-tail classification method of \cite{kang2019decoupling} to relieve the bottleneck of RL and proposed sFT and wFT.
Both methods were carefully designed for RL models, but these were not the only way to employ long-tail classification methods.
In this section, we discuss the other possible adaptations based on \cite{raunak-etal-2020-long}.

\cite{raunak-etal-2020-long} explored ways to employ long-tail classification methods for machine translation.
Their first method was $\tau$-normalization ($\tau$-norm), which directly adopted the method of \cite{kang2019decoupling}.
Based on an observation that the norm of classifier parameters correlates with the frequency of the classes, they normalized the classifier weight $\bm{W}$ as follows:
\begin{align}
    \label{eq:tau}
    \widetilde{\bm{W}}_{w_i} &= \frac{\bm{W}_{w_i}}{\| \bm{W}_{w_i} \|^{\tau}},
\end{align}
where $\bm{W}_{w_i} \in \mathbb{R}^{d}$ indicates a vector at the index of a word $w_i$ and $\tau$ is a temperature hyperparameter that controls the degree of the normalization.

The other methods of \cite{raunak-etal-2020-long} were Focal loss (FL) and Anti-Focal loss (AFL).
AFL is a variant of FL~\cite{lin2017focal}, which was aimed at reweighting the loss according to the confidence of the model predictions.
Let $p_{\theta}^t = p_{\theta}(w^g_t \mid w^g_{<t}, I)$.
FL and AFL in image captioning are then written as follows:
\begin{align}
    \label{eq:focal}
    \mathcal{L}_{\mathrm{FL}}(\theta) &= -\frac{1}{T} \sum^{T}_{t=1}{(1 - p_{\theta}^t)^{\gamma}\log p_{\theta}^t}, \\
    \label{eq:antifocal}
    \mathcal{L}_{\mathrm{AFL}}(\theta) &= -\frac{1}{T} \sum^{T}_{t=1}{(1 + \alpha p_{\theta}^t)^{\gamma}\log p_{\theta}^t},
\end{align}
where $\gamma$ and $\alpha$ are hyperparamters that control the degree of the reweighting.
Other work also explored ways to employ long-tail classification methods for text generation, but those approaches are categorized as either $\tau$-norm~\cite{nguyen2018improving} or variants of FL~\cite{gu2020token,jiang2019improving,wu2020importance}, which we already explored above.

\begin{table*}[t]
\centering
\scalebox{0.70}{
\begin{tabular}{lcccccccccccc}  
\toprule
 & \multicolumn{3}{c}{\emph{Vocabulary}} & \multicolumn{6}{c}{\emph{Standard Evaluation}} & \multicolumn{3}{c}{\emph{Discriminativeness}} \\
\cmidrule(lr){2-4}
\cmidrule(lr){5-10}
\cmidrule(lr){11-13}
 & Unique-1 & Unique-S & Length & CIDEr & SPICE & BERTS+ & TIGEr & CLIPS & RefCLIPS & R@1 & R@5 & R@10 \\
\midrule
\textbf{Att2in RL} & 445 & 2,524 & 9.3 & \textbf{117.4} & \textbf{20.5} & 43.6 & 73.9 & 73.0 & 79.7 & 16.3 & 41.9 & 57.2 \\
\rowcolor[rgb]{0.9, 0.9, 0.9}
\ \ + sFT & 880 & 3,156 & 9.0 & 115.4 & 20.4 & \textbf{43.9} & 74.3 & 73.7 & \textbf{80.3} & 20.1 & 48.0 & 62.8 \\
\rowcolor[rgb]{0.9, 0.9, 0.9}
\ \ + wFT & \textbf{1,197} & \textbf{3,732} & 8.9 & 104.3 & 19.5 & 43.1 & 74.2 & 73.9 & 80.2 & 20.6 & 49.7 & 64.5 \\
\rowcolor[rgb]{0.9, 0.9, 0.9}
\ \ + wFT (BP decoding) & 1,102 & 3,615 & 9.4 & 109.3 & 20.1 & 43.7 & \textbf{74.4} & \textbf{74.0} & 80.2 & \textbf{21.1} & \textbf{50.5} & \textbf{64.8} \\
\ \ + $\tau$-norm & 437 & 2,414 & 9.1 & 117.3 & 20.4 & 43.5 & 73.8 & 72.9 & 79.7 & 15.4 & 40.7 & 55.8 \\
\ \ + FL & 903 & 3,217 & 9.0 & 114.8 & 20.4 & 43.8 & 74.3 & 73.7 & \textbf{80.3} & 20.1 & 48.1 & 63.2 \\
\ \ + AFL & 886 & 3,116 & 9.0 & 115.3 & 20.4 & 43.8 & 74.3 & 73.7 & \textbf{80.3} & 19.7 & 47.6 & 62.7 \\
\ \ + Nucleus sampling & 475 & 2,726 & 9.3 & 116.5 & 20.3 & 43.5 & 73.9 & 72.9 & 79.7 & 16.5 & 41.9 & 57.1 \\
\midrule
\textbf{UpDown RL} & 577 & 3,103 & 9.5 & \textbf{122.7} & \textbf{21.5} & \textbf{44.2} & 74.6 & 74.0 & 80.5 & 21.1 & 49.9 & 64.6 \\
\rowcolor[rgb]{0.9, 0.9, 0.9}
\ \ + sFT & 1,190 & 3,788 & 9.2 & 115.9 & 21.0 & \textbf{44.2} & \textbf{74.9} & 74.8 & \textbf{80.9} & 25.0 & 56.8 & 71.2 \\
\rowcolor[rgb]{0.9, 0.9, 0.9}
\ \ + wFT & \textbf{1,479} & \textbf{4,268} & 9.1 & 101.8 & 19.5 & 43.1 & 74.6 & 74.9 & 80.7 & 26.0 & 57.6 & 72.2 \\
\rowcolor[rgb]{0.9, 0.9, 0.9}
\ \ + wFT (BP decoding) & 1,275 & 4,177 & 9.6 & 110.0 & 20.6 & 44.1 & \textbf{74.9} & \textbf{75.0} & 80.8 & \textbf{26.7} & \textbf{58.7} & \textbf{72.4} \\
\ \ + $\tau$-norm & 576 & 2,967 & 9.3 & 122.6 & 21.3 & \textbf{44.2} & 74.4 & 73.8 & 80.5 & 19.6 & 48.1 & 63.4 \\
\ \ + FL & 1,201 & 3,830 & 9.2 & 114.9 & 20.9 & 44.1 & \textbf{74.9} & 74.7 & \textbf{80.9} & 25.2 & 57.0 & 70.9 \\
\ \ + AFL & 1,171 & 3,760 & 9.2 & 116.4 & 20.9 & \textbf{44.2} & \textbf{74.9} & 74.7 & \textbf{80.9} & 24.9 & 56.6 & 70.7 \\
\ \ + Nucleus sampling & 592 & 3,339 & 9.5 & 120.7 & 21.3 & \textbf{44.2} & 74.6 & 73.9 & 80.4 & 20.9 & 49.7 & 64.4 \\
\midrule
\textbf{Transformer RL} & 753 & 3,433 & 9.2 & \textbf{127.7} & \textbf{22.5} & \textbf{45.1} & 75.0 & 75.0 & 81.3 & 26.6 & 56.2 & 70.5 \\
\rowcolor[rgb]{0.9, 0.9, 0.9}
\ \ + sFT & 1,458 & 3,959 & 9.1 & 118.7 & 21.7 & 44.8 & \textbf{75.2} & 75.6 & 81.5 & 30.6 & 62.3 & 75.7 \\
\rowcolor[rgb]{0.9, 0.9, 0.9}
\ \ + wFT & 1,776 & 4,274 & 9.1 & 103.1 & 20.0 & 43.3 & 74.8 & 75.8 & 81.2 & 32.5 & 64.5 & 77.1 \\
\rowcolor[rgb]{0.9, 0.9, 0.9}
\ \ + wFT (BP decoding) & \textbf{1,964} & \textbf{4,373} & 9.4 & 107.3 & 21.1 & 44.2 & \textbf{75.2} & \textbf{76.1} & 81.5 & \textbf{33.5} & \textbf{65.9} & \textbf{78.2} \\
\ \ + $\tau$-norm & 1,027 & 3,483 & 9.2 & 124.4 & 22.1 & 44.9 & 74.8 & 74.9 & 81.2 & 26.1 & 55.8 & 69.7 \\
\ \ + FL & 1,523 & 4,018 & 9.1 & 116.5 & 21.4 & 44.6 & \textbf{75.2} & 75.7 & 81.5 & 31.2 & 63.1 & 76.3 \\
\ \ + AFL & 1,402 & 3,908 & 9.1 & 120.5 & 21.9 & 44.8 & \textbf{75.2} & 75.6 & \textbf{81.6} & 30.0 & 62.1 & 75.9 \\
\ \ + Nucleus sampling & 1,053 & 3,751 & 9.3 & 123.7 & 22.0 & 44.8 & 74.9 & 75.0 & 81.2 & 26.9 & 55.8 & 70.4 \\
\bottomrule
\end{tabular}
}
\caption{
Comparison with the other long-tail
classification methods.
Automatic evaluation results on the MS~COCO test set.
\emph{Unique-1} and \emph{Unique-S} indicate the number of unique unigrams and sentences, respectively.
\emph{Length} is the average length of output captions.
}
\label{tab:long}
\end{table*}

We compared our methods (sFT and wFT) with $\tau$-norm, FL, and AFL.
In our experiments, we normalized the bias term $\bm{b}$\footnote{$\tilde{\bm{b}}=\frac{\bm{b}}{\| \bm{b} \|^{\tau}}$, where the value of the hyperparameter $\tau$ was set to the same as that of $\widetilde{\bm{W}}$.} in addition to the weight term $\bm{W}$ as we found it performed better than normalizing the weight term only.
We applied FL and AFL as the alternative weighting to BP for a fair comparison with our methods.
That is, we fine-tuned the classifier parameters by optimizing $\mathcal{L}_{\mathrm{FL}}(\hat{\theta})$ or $\mathcal{L}_{\mathrm{AFL}}(\hat{\theta})$, where $\hat{\theta}$ were initialized with the pre-trained RL models.
We used the best hyperparameters reported in \cite{raunak-etal-2020-long}: $\tau=0.2$, $\gamma=1$, and $\alpha=1$.
Similar to our models, other hyperparameters were set to the same values as the baseline models, except for the epoch size and learning rate.
We explored the same values for these hyperparameters as our models: we set the epoch size for fine-tuning to 1 and searched for the best learning rates from $\{1\textrm{e-}3,1\textrm{e-}4,1\textrm{e-}5,1\textrm{e-}6\}$.
We selected the best learning rate according to the R@1 scores in the validation set.
The best learning rate was 1e-4 for Att2in RL + FL/AFL, 1e-4 for UpDown RL + FL/AFL, and 1e-5 for Transformer RL + FL/AFL.
Note that we did not explore the learning rate for $\tau$-norm because it does not require training.

In open-ended text generation tasks, \eg, story generation and text generation after prompts, stochastic sampling methods are used instead of beam search to increase the diversity in output text~\cite{holtzman2020curious,basu2020mirostat,meister2022typical}.
Although image captioning does not fall in the category of open-ended text generation as input images tightly scope the correctness of captions, we additionally test whether the randomness in stochastic sampling can increase the output vocabulary.
We used Nucleus sampling~\cite{holtzman2020curious} with a hyperparameter $p=0.95$, which is the best hyperparameter reported~\cite{holtzman2020curious,meister2022typical}.

Table~\ref{tab:long} shows the results.
$\tau$-norm and Nucleus sampling showed the similar results.
Both methods slightly increased the output vocabulary but the performance generally remained the same as the baseline models.
These results indicate that the output vocabulary cannot be significantly increased while maintaining the relative probability of words: Nucleus sampling samples according to the original output distributions and $\tau$-norm changes the distribution only by the difference in the norm, basically flattening the distribution.
In contrast, FL and AFL drastically change the relative probability of words by refining the mapping from encoded features to low-frequency words, as with sFT and wFT.
They successfully increased the vocabulary size and discriminativeness.
However, the gains were smaller than those of wFT.

\begin{figure}[t]
    \centering
    \includegraphics[width=1.0\columnwidth,keepaspectratio]{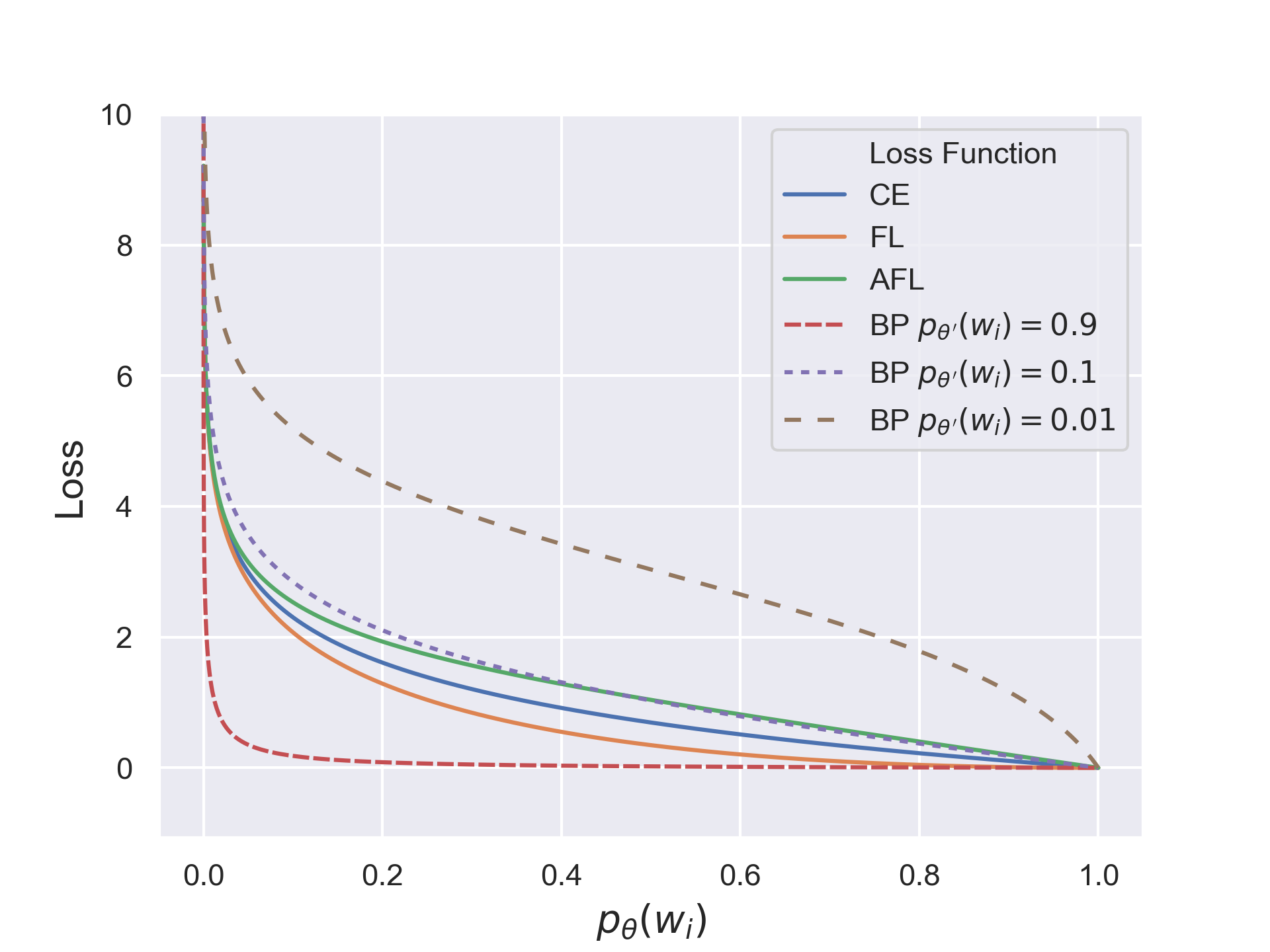}
    \caption{
    Visualization of the losses: CE $-\log p_{\theta} (w_i)$, BP $-\log p_{\theta, \theta'}(w_i)$, FL $(1 - p_{\theta}(w_i))^{\gamma} \log p_{\theta}(w_i)$, and AFL $(1 + \alpha p_{\theta}(w_i))^{\gamma} \log p_{\theta}(w_i)$.
    We set $\beta=\beta'=1$, $\gamma=1$, and $\alpha=1$.
    }
    \label{fig:focal}
\end{figure}

To analyze the cause of the difference between FL, AFL, and the BP loss (wFT), we visualized the losses in Figure~\ref{fig:focal}.
FL suppresses the loss when a model is confident, whereas AFL increases the loss when a model is moderately confident.
Compared with these losses, BP changes the loss more drastically.
When the frequency-biased policy $p_{\theta'}$ is highly confident, BP strictly suppresses the loss to prevent further learning on that word; when $p_{\theta'}$ is not confident, BP highly increases the loss to encourage the learning on that word.
This drastic rebalancing of the loss resulted in wFT's larger vocabulary size and higher discriminativeness.

\begin{table*}[t]
\centering
\scalebox{0.70}{
\begin{tabular}{lcccccccccccc}  
\toprule
 & \multicolumn{3}{c}{\emph{Vocabulary}} & \multicolumn{6}{c}{\emph{Standard Evaluation}} & \multicolumn{3}{c}{\emph{Discriminativeness}} \\
\cmidrule(lr){2-4}
\cmidrule(lr){5-10}
\cmidrule(lr){11-13}
 & Unique-1 & Unique-S & Length & CIDEr & SPICE & BERTS+ & TIGEr & CLIPS & RefCLIPS & R@1 & R@5 & R@10 \\
\midrule
\textbf{Att2in RL} & 435 & 2,583 & 9.3 & \textbf{116.5} & \textbf{20.3} & 43.6 & N/A & 73.1 & 79.8 & 16.2 & 42.5 & 57.0 \\
\rowcolor[rgb]{0.9, 0.9, 0.9}
\ \ + sFT & 874 & 3,189 & 9.0 & 113.7 & 20.1 & \textbf{43.7} & N/A & 73.9 & \textbf{80.3} & 19.2 & 47.9 & 62.9 \\
\rowcolor[rgb]{0.9, 0.9, 0.9}
\ \ + wFT & \textbf{1,196} & \textbf{3,792} & 9.0 & 104.8 & 19.3 & 43.2 & N/A & \textbf{74.2} & \textbf{80.3} & 19.6 & 50.4 & 64.6 \\
\rowcolor[rgb]{0.9, 0.9, 0.9}
\ \ + wFT (BP decoding) & 1,105 & 3,633 & 9.4 & 108.6 & 20.0 & \textbf{43.7} & N/A & 74.1 & \textbf{80.3} & \textbf{20.6} & \textbf{50.6} & \textbf{64.9} \\
\midrule
\textbf{UpDown RL} & 563 & 3,161 & 9.5 & \textbf{122.3} & \textbf{21.3} & \textbf{44.2} & N/A & 74.2 & 80.6 & 20.6 & 50.2 & 65.7 \\
\rowcolor[rgb]{0.9, 0.9, 0.9}
\ \ + sFT & 1,222 & 3,805 & 9.2 & 115.3 & 20.7 & 44.1 & N/A & 74.9 & \textbf{80.9} & 24.6 & 56.2 & 70.9 \\
\rowcolor[rgb]{0.9, 0.9, 0.9}
\ \ + wFT & \textbf{1,502} & \textbf{4,301} & 9.1 & 100.5 & 19.2 & 43.0 & N/A & 75.0 & 80.7 & 26.1 & 57.4 & 71.4 \\
\rowcolor[rgb]{0.9, 0.9, 0.9}
\ \ + wFT (BP decoding) & 1,278 & 4,226 & 9.6 & 108.9 & 20.5 & 43.9 & N/A & \textbf{75.1} & \textbf{80.9} & \textbf{26.4} & \textbf{58.6} & \textbf{73.6} \\
\midrule
\textbf{Transformer RL} & 713 & 3,432 & 9.2 & \textbf{126.4} & \textbf{22.1} & \textbf{45.0} & N/A & 75.0 & 81.2 & 25.4 & 56.3 & 69.8 \\
\rowcolor[rgb]{0.9, 0.9, 0.9}
\ \ + sFT & 1,496 & 3,953 & 9.1 & 118.4 & 21.4 & 44.6 & N/A & 75.7 & \textbf{81.5} & 30.2 & 62.7 & 75.8 \\
\rowcolor[rgb]{0.9, 0.9, 0.9}
\ \ + wFT & 1,836 & 4,268 & 9.1 & 102.2 & 19.8 & 43.2 & N/A & 75.9 & 81.3 & 32.2 & 64.3 & 76.8 \\
\rowcolor[rgb]{0.9, 0.9, 0.9}
\ \ + wFT (BP decoding) & \textbf{2,004} & \textbf{4,392} & 9.4 & 105.6 & 20.6 & 43.9 & N/A & \textbf{76.1} & 81.4 & \textbf{32.8} & \textbf{66.1} & \textbf{79.0} \\
\bottomrule
\end{tabular}
}
\caption{
Automatic evaluation results on the MS~COCO \emph{validation} set.
\emph{Unique-1} and \emph{Unique-S} indicate the number of unique unigrams and sentences, respectively.
\emph{Length} is the average length of output captions.
TIGEr scores are N/A as the TIGEr evaluation tool currently does not support evaluation on the MS~COCO validation set.
}
\label{tab:validation}
\end{table*}

\section{Validation Performance for Reproduction}
\label{appendix:validation}
Table~\ref{tab:validation} shows the performance of our models on the MS~COCO validation set.
We report these results for the future reproduction of our experiments.
The code is available at \url{https://github.com/ukyh/switch_disc_caption.git}.

\begin{table*}[t]
\centering
\scalebox{0.70}{
\begin{tabular}{lcccccccccccc}  
\toprule
 & \multicolumn{3}{c}{\emph{Vocabulary}} & \multicolumn{6}{c}{\emph{Standard Evaluation}} & \multicolumn{3}{c}{\emph{Discriminativeness}} \\
\cmidrule(lr){2-4}
\cmidrule(lr){5-10}
\cmidrule(lr){11-13}
 & Unique-1 & Unique-S & Length & CIDEr & SPICE & BERTS+ & TIGEr & CLIPS & RefCLIPS & R@1 & R@5 & R@10 \\
\midrule
\textbf{VinVL RL} & 1,126 & 4,298 & 10.0 & \textbf{140.9} & \textbf{25.2} & \textbf{46.1} & \textbf{75.7} & 77.6 & \textbf{83.3} & 36.1 & 68.5 & 80.2 \\
\rowcolor[rgb]{0.9, 0.9, 0.9}
\ \ + sFT & 1,834 & 4,649 & 10.0 & 126.0 & 23.8 & 45.5 & 75.6 & \textbf{78.2} & \textbf{83.3} & 39.2 & \textbf{72.1} & 83.8 \\
\rowcolor[rgb]{0.9, 0.9, 0.9}
\ \ + wFT & \textbf{1,852} & 4,652 & 10.0 & 124.9 & 23.7 & 45.5 & 75.6 & \textbf{78.2} & \textbf{83.3} & 39.2 & 72.0 & 83.9  \\
\rowcolor[rgb]{0.9, 0.9, 0.9}
\ \ + wFT (BP decoding) & 1,734 & \textbf{4,717} & 9.8 & 122.4 & 23.5 & 45.2 & \textbf{75.7} & \textbf{78.2} & \textbf{83.3} & \textbf{39.6} & \textbf{72.1} & \textbf{84.6} \\
\bottomrule
\end{tabular}
}
\caption{
Test on the more recent captioning model.
Automatic evaluation results on the MS~COCO test set.
\emph{Unique-1} and \emph{Unique-S} indicate the number of unique unigrams and sentences, respectively.
\emph{Length} is the average length of output captions.
}
\label{tab:vinvl}
\end{table*}

\section{Effectiveness on More Recent Models}
\label{appendix:recent models}
To further demonstrate the effectiveness of our methods, we tested our fine-tuning methods on a more recent captioning model, \textbf{VinVL}~\cite{zhang2021vinvl,li2020oscar}.
VinVL boosts its performance through large-scale cross-modal pre-training.
The significant performance improvements have made VinVL a popular captioning model and one of the most advanced captioning models available today~\cite{stefanini2021show,wang2022ofa,nguyen2022grit}.

We used the best-performing pre-trained model as our baseline: \texttt{coco\_captioning\_large\_scst} model that is publicly available at \url{https://github.com/microsoft/Oscar/blob/master/VinVL_MODEL_ZOO.md#Image-Captioning-on-COCO}.
Note that this model was trained with the standard RL~\cite{rennie2017}.

As in the previous experiments, we applied our fine-tuning methods for one epoch only.
We searched for the best learning rates for fine-tuning from $\{1\textrm{e-}5,1\textrm{e-}6\}$, and the inverse-temperature hyperparameter $\beta'$ of Eq.~(\textcolor{red}{7}) from $\{0.01, 0.1, 1\}$.
Other hyperparameters were set to the same as the baseline model.
The best learning rate was 1e-5.
The best $\beta'$ was 0.01 for wFT with $p_{\theta}$ decoding and 1 for wFT with BP decoding.
Note that sFT does not use $\beta'$.

Table~\ref{tab:vinvl} shows similar results as Table \textcolor{red}{1} \emph{in the main paper}.
Our methods significantly increased the vocabulary size from the baseline and accordingly enhanced the discriminativeness.
The standard evaluation metrics also showed the same tendency.
Although our models scored lower than the baseline in the conventional exact-matching metrics (CIDEr and SPICE), the gap became smaller in the more advanced soft-matching metrics (BERTS+ and TIGEr).
In the state-of-the-art soft-matching metrics (CLIPS and RefCLIPS), our models achieved the same or even higher scores than the baseline.
These results show that our methods are also effective on the more recent model.
Moreover, these results further validate that our methods can switch any off-the-shelf RL models to discriminativeness-aware models while maintaining the overall quality of captions.

\begin{table*}[t]
\centering
\scalebox{0.70}{
\begin{tabular}{lcccccccccccc}  
\toprule
 & \multicolumn{3}{c}{\emph{Vocabulary}} & \multicolumn{6}{c}{\emph{Standard Evaluation}} & \multicolumn{3}{c}{\emph{Discriminativeness}} \\
\cmidrule(lr){2-4}
\cmidrule(lr){5-10}
\cmidrule(lr){11-13}
 & Unique-1 & Unique-S & Length & CIDEr & SPICE & BERTS+ & TIGEr & CLIPS & RefCLIPS & R@1 & R@5 & R@10 \\
\midrule
\textbf{Transformer* RL (CIDEr)} & 691 & 3,650 & 9.5 & \textbf{126.0} & \textbf{22.8} & \textbf{45.2} & 74.6 & 75.8 & 81.6 & 27.1 & 57.2 & 70.6 \\
\rowcolor[rgb]{0.9, 0.9, 0.9}
\ \ + sFT & 1,265 & 4,071 & 9.1 & 122.9 & 22.2 & \textbf{45.2} & \textbf{74.8} & 76.4 & \textbf{82.0} & 31.4 & 62.0 & 75.0 \\
\rowcolor[rgb]{0.9, 0.9, 0.9}
\ \ + wFT & \textbf{1,546} & 4,337 & 9.0 & 111.3 & 21.0 & 44.2 & 74.5 & 76.5 & 81.8 & 31.6 & 63.3 & 75.7 \\
\rowcolor[rgb]{0.9, 0.9, 0.9}
\ \ + wFT (BP decoding) & 1,543 & \textbf{4,471} & 9.5 & 112.3 & 21.7 & 44.9 & \textbf{74.8} & \textbf{76.9} & 81.9 & \textbf{34.0} & \textbf{65.4} & \textbf{78.4} \\
\midrule
\textbf{Transformer* RL (CLIPS + Grammar)} & 952 & 4,847 & 13.0 & 74.1 & 19.8 & 43.6 & \textbf{75.0} & \textbf{79.2} & 81.2 & 44.2 & 77.0 & 86.9 \\
\rowcolor[rgb]{0.9, 0.9, 0.9}
\ \ + sFT & 969 & 4,848 & 12.8 & 76.4 & 20.1 & 43.8 & \textbf{75.0} & \textbf{79.2} & 81.2 & 44.6 & \textbf{77.3} & 87.0 \\
\rowcolor[rgb]{0.9, 0.9, 0.9}
\ \ + wFT & 969 & 4,847 & 12.9 & 76.4 & 20.1 & 43.8 & \textbf{75.0} & \textbf{79.2} & 81.2 & 44.8 & 77.2 & \textbf{87.1}  \\
\rowcolor[rgb]{0.9, 0.9, 0.9}
\ \ + wFT (BP decoding) & \textbf{1,001} & \textbf{4,853} & 12.2 & \textbf{82.5} & \textbf{20.6} & \textbf{44.1} & \textbf{75.0} & \textbf{79.2} & \textbf{81.3} & \textbf{45.5} & 77.2 & \textbf{87.1} \\
\midrule
\textbf{Transformer* Only CE} & 1,174 & 3,637 & 9.4 & 113.8 & 20.9 & 44.1 & 74.0 & 75.1 & 81.1 & 26.2 & 55.2 & 68.6 \\
\bottomrule
\end{tabular}
}
\caption{
Test on the more recent discriminativeness-aware model.
Transformer* used a different image encoder than the other transformer models tested in this paper.
Automatic evaluation results on the MS~COCO test set.
\emph{Unique-1} and \emph{Unique-S} indicate the number of unique unigrams and sentences, respectively.
\emph{Length} is the average length of output captions.
}
\label{tab:clips}
\end{table*}

\section{Comparison and Combination with More Recent Discriminativeness-Aware Models}
\label{appendix:recent disc models}
Contemporaneous to our work, \cite{cho-etal-2022-fine} showed that maximizing reference-free CLIPS-based reward enhanced discriminativeness significantly.
In this section, we clarify the advantages of our methods over the CLIPS-based RL by comparing and combining our methods with it.

The pre-trained models of \cite{cho-etal-2022-fine} are publicly available at \url{https://github.com/j-min/CLIP-Caption-Reward}.
We used the transformer model trained with the standard CIDEr reward (\textbf{Transformer* RL (CIDEr)}; \texttt{clipRN50\_cider}) and the one trained with the reward proposed by \cite{cho-etal-2022-fine} (\textbf{Transformer* RL (CLIPS + Grammar)}; \texttt{clipRN50\_clips\_grammar})\footnote{Note that \texttt{clipRN50} does not mean that the model used the CLIPS-based reward. It denotes that the model used CLIP~\cite{pmlr-v139-radford21a} as the image encoder, unlike the other models tested in this paper.}.
The proposed reward is computed by the weighted sum of CLIPS and grammaticality scores.
We also included \textbf{Transformer* Only CE} (\texttt{clipRN50\_mle}) in the comparison as the baseline without RL.

As in the previous experiments, we applied our fine-tuning methods for one epoch only.
We searched for the best learning rates for fine-tuning from $\{1\textrm{e-}5,1\textrm{e-}6,1\textrm{e-}7\}$, and the inverse-temperature hyperparameter $\beta'$ of Eq.~(\textcolor{red}{7}) from $\{0.01, 0.1, 1\}$.
Other hyperparameters were set to the same as the baseline model.
The best learning rate for Transformer* RL (CIDEr) was 1e-5; the best $\beta'$ was 0.1 for wFT with $p_{\theta}$ decoding and 1 for wFT with BP decoding.
The best learning rates for Transformer* RL (CLIPS + Grammar) were 1e-6 for wFT with BP decoding and 1e-7 for the others; the best $\beta'$ was 1 for wFT with both decoding methods.
Note that sFT does not use $\beta'$.

Table~\ref{tab:clips} shows the results.
Similar to the previous results, our methods significantly enhanced the vocabulary size and discriminativeness from the RL models while maintaining or even increasing the scores in the state-of-the-art soft-matching metrics.
The CLIPS~+~Grammar reward also achieved the high discriminativeness compared with the standard CIDEr reward.

However, the improvement of the CLIPS-based RL came at the expense of the \emph{conciseness} and overall quality of captions in contrast to our methods: compared to Transformer* RL (CIDEr), Transformer* RL (CLIPS + Grammar) significantly increased the sentence length and decreased scores in the standard evaluation metrics, including the current best-performing metric, RefCLIPS.
Although increasing the sentence length is one way to describe images in detail, concise description is more desirable to convey the most characteristic information clearly and efficiently~\cite{sadovnik2012image}.
Despite the longer sentence length, the side effect was still observed: CLIPS-based RL decreased the output vocabulary from the Only CE baseline.

These results indicate that our methods and the CLIPS-based RL increased discriminativeness by different factors: more specific vocabulary and longer descriptions, respectively.
In other words, the contribution of our methods is orthogonal to that of the CLIPS-based RL.
To utilize the strength of each, we applied our methods to the CLIPS-based RL model.
Although the CLIPS-based RL achieved the high discriminativeness and relatively large vocabulary size due to the longer sentences, our methods further enhanced the discriminativeness and vocabulary size.
Surprisingly, our methods also improved the standard evaluation scores, including exact-matching scores.
This result suggests that our fine-tuning with ground-truth captions restored the overall quality of captions, which was degraded by over-optimization for reference-free CLIPS.

Another critical advantage of our methods is computational efficiency.
Training of CLIPS-based RL took \emph{one day using eight GPUs}~\cite{cho-etal-2022-fine}, while ours only took \emph{40 minutes using a single GPU}.

The above results conclude that our methods are orthogonal to the more recent discriminative image captioning method and have important advantages in conciseness and efficiency.

{\small
\bibliographystyle{ieee_fullname}
\bibliography{wacv2023supp}
}